\documentclass{article}

\PassOptionsToPackage{numbers, compress}{natbib}

\usepackage[preprint]{neurips_2026}

\usepackage[utf8]{inputenc} 
\usepackage[T1]{fontenc}    
\usepackage{hyperref}       
\hypersetup{
  colorlinks=true,
  linkcolor=blue,
  citecolor=blue,
  urlcolor=blue,
}
\usepackage{url}            
\usepackage{booktabs}       
\usepackage{amsfonts}       
\usepackage{nicefrac}       
\usepackage{microtype}      
\usepackage{xcolor}         
\usepackage{multirow}
\usepackage{graphicx}
\usepackage{subcaption}
\usepackage{float}
\usepackage{colortbl}       
\usepackage{paralist}
\usepackage{tabularx}
\usepackage{array}
\usepackage{pdflscape}
\usepackage{placeins}
\usepackage{etoc}

\newcolumntype{Y}{>{\raggedright\arraybackslash}X}

\usepackage{amsmath}
\usepackage{amssymb}
\usepackage{mathtools}
\usepackage{amsthm}
\usepackage{algorithm}
\usepackage{algorithmic}
\usepackage{fontawesome5} 

\definecolor{projectpink}{HTML}{FF00A8}

\newcommand{\projecthref}[2]{%
  \href{#1}{\textcolor{projectpink}{#2}}%
}

\usepackage[capitalize,noabbrev]{cleveref}

\theoremstyle{plain}

\theoremstyle{definition}

\theoremstyle{remark}

\usepackage[textsize=tiny]{todonotes}

\title{AutoRubric-T2I: Robust Rule-Based Reward Model for Text-to-Image Alignment}

\author{%
\begin{tabular}{c}
\textbf{Kuei-Chun Kao}\textsuperscript{1}
\quad
\textbf{Daixuan Huo}\textsuperscript{1}
\quad
\textbf{Yuanhao Ban}\textsuperscript{1,2}
\quad
\textbf{Cho-Jui Hsieh}\textsuperscript{1,2}
\\
\textsuperscript{1}University of California, Los Angeles
\quad
\textsuperscript{2}Arena
\\
\texttt{johnson0213@g.ucla.edu, chohsieh@cs.ucla.edu}
\\[0.4em]
\textbf{Project Page: }
\projecthref{https://johnsonkao0213.github.io/AutoRubric-T2I}{\faGlobe\ AutoRubric-T2I}
\quad
\projecthref{https://github.com/johnsonkao0213/AutoRubric-T2I}{\faGithub\ Code}
\end{tabular}
}

\begin{document}

\maketitle

\etocdepthtag.toc{main}
\begin{abstract}
Aligning Text-to-Image (T2I) generation models with human preferences increasingly relies on image reward models that score or rank generated images according to prompt alignment and perceptual quality. Existing reward models are commonly trained as Bradley-Terry (BT) preference models on large-scale human preference corpora, making them costly to train, difficult to adapt, and opaque in their evaluation criteria. Meanwhile, Vision-Language Model (VLM) judges can provide more fine-grained assessments through textual rubrics, but their manually designed or heuristically generated scoring rules may fail to reliably reflect human preferences.
In this paper, we propose \textbf{AutoRubric-T2I}, the first rubric learning framework in T2I that automatically synthesizes and selects explicit rubrics for guiding VLM judges. AutoRubric-T2I first synthesizes reasoning traces from preference pairs into candidate rubrics, then uses a VLM judge to score paired images under each rubric, producing pairwise rubric-score differences for preference learning. To remove noisy and redundant rules, we further employ a \textbf{$\ell_1$-Regularized Logistic Regression Refiner}, which selects the Top-$N$ most discriminative rubrics.
Extensive evaluations show that AutoRubric-T2I produces high-quality, interpretable reward signals using less than 0.01\% of the annotated preference data, substantially reducing the need for large-scale reward-model training. On image reward benchmarks such as MMRB2, AutoRubric-T2I outperforms strong reward model baselines. We further validate AutoRubric-T2I as an RL reward on downstream T2I tasks, including TIIF and UniGenBench++, where it improves generation quality over scalar reward models using the Flow-GRPO pipeline on diffusion models.
\end{abstract}
\vspace{-6mm}
\section{Introduction}
\vspace{-2mm}
Recent advancements in T2I generation have made human preference alignment a central objective. Image reward models provide a practical mechanism for this alignment by learning to predict human judgments over generated images. In the T2I setting, an image reward model typically takes a text prompt together with one or more generated images as input and outputs either scalar reward scores or pairwise preferences indicating which image better satisfies the prompt and human quality expectations. These models are widely used for candidate ranking, best-of-$N$ selection, data filtering, reinforcement fine-tuning (RFT), and automatic evaluation of T2I systems~\cite{jiang2024genai,liu2025flow,jiang2025t2i}.

Existing image reward models mainly fall into two categories. The first category consists of learned BT preference models trained on large-scale human preference corpora, such as ImageReward~\cite{xu2023imagereward}, PickScore~\cite{kirstain2023pick}, HPSv2~\cite{wu2023hpsv2}, and HPSv3~\cite{ma2025hpsv3}. These models can capture real-world human preferences and are effective for global image ranking, but they require massive annotated datasets and expensive fine-tuning. Moreover, because they usually compress multiple evaluation dimensions into a single scalar score, they provide limited transparency and may overlook fine-grained visual errors such as incorrect object counts, missing attributes, distorted anatomy, or violated spatial relations~\cite{hu2023tifa,li2024genaibench,hong2026understanding}. The second category consists of VLM-based judges and question-answering-based evaluators, which evaluate images through textual prompts, visual questions, or rubrics~\cite{hu2023tifa,li2024genaibench,hu2025multimodal,wang2025unified}. These judges can assess fine-grained visual correctness when properly instructed, but their criteria are typically manually specified or heuristically generated rather than learned from human preferences. As a result, their judgments may not reliably correlate with actual human preference; for example, prompted VLM judges can be up to 10\% worse than learned BT reward models on HPS or PickScore preference datasets in Table~\ref{tab:full_mmrb2_results}.

These limitations motivate \textbf{rubric-based reward modeling}, which replaces implicit scalar rewards with multi-dimensional evaluation criteria~\cite{xie2025autorubric,wang2025autorule,gunjal2025rubrics,zhang2026chasing,liu2025openrubrics}. Rubrics make the reward signal more interpretable by decomposing human preferences into explicit rules. Recent works have explored rubrics as reward signals for LLM alignment and post-training~\cite{xie2025autorubric,wang2025autorule,gunjal2025rubrics,huang2025reinforcement,xu2026alternating,li2026rubrichub}, and emerging T2I work has begun to study rubric rewards for image generation~\cite{feng2025rubricrl}. However, existing rubric-based approaches often rely on manually designed or heuristically generated rubrics, leaving open the question of how to automatically derive, select, and refine rubrics that better align with human preferences.

To address this gap, we propose \textbf{AutoRubric-T2I}, the first rubric learning framework that automatically derives and refines an explicit rubric set for guiding off-the-shelf VLM judges in T2I reward modeling. Instead of fine-tuning a reward model, AutoRubric-T2I learns which rubrics are most predictive of human preferences and iteratively improves them through failure analysis. This design preserves the interpretability of rubric-based evaluation while avoiding the cost and opacity of training a dense scalar reward model.

To achieve this, we formulate automated rubric learning as a sparse logistic regression problem within an infinite-dimensional space. We then introduce an iterative block coordinate descent method that dynamically adds new coordinates to the working set and employs $\ell_1$-regularization to assign weights and prune redundant coordinates (rubrics). This formulation is related to sparse function approximation and coordinate-selection methods such as orthogonal matching pursuit and sparse random features~\cite{pati1993orthogonal,jain2011orthogonal,yen2014sparse}. To enhance efficiency, we integrate a hard-pair mining algorithm for rubric refinement, ensuring that only the most informative coordinates are prioritized during the learning process.

Our main contributions are as follows:
\begin{compactitem}
    \item \textbf{Sparse Rubric Learning for T2I Reward Modeling:} We introduce AutoRubric-T2I, a framework that learns a compact, weighted set of natural-language rubrics from image preference data, enabling interpretable VLM-based reward modeling without fine-tuning.

    \item \textbf{Failure-Driven Rubric Refinement:} We formulate rubric selection as an $\ell_1$-regularized logistic regression problem over VLM-scored rubric features and iteratively expand the rubric pool through curriculum-bucketed hard-pair mining.

    \item \textbf{Strong Preference Prediction and Downstream Alignment:} AutoRubric-T2I achieves strong preference prediction on MMRB2 among open-source reward models and improves downstream RFT on T2I tasks such as TIIF and UniGenBench++ using Flow-GRPO.
\end{compactitem}
\vspace{-4mm}

\section{Related Work}
\vspace{-2mm}
\subsection{Text-to-Image Preference Alignment and Reward Modeling}
\vspace{-1mm}
Aligning text-to-image (T2I) models with human preferences commonly relies on reward models trained from human preference data. Many image reward models are trained from pairwise comparisons, often with a Bradley-Terry style objective, but are deployed as pointwise scorers that assign a scalar reward to each prompt-image pair. Large-scale preference datasets and models such as PickScore~\cite{kirstain2023pick} enabled automatic ranking of generated images according to human judgments. Subsequent reward models, including ImageReward~\cite{xu2023imagereward}, HPSv2~\cite{wu2023hpsv2}, and HPSv3~\cite{ma2025hpsv3}, further improved visual preference modeling by capturing visual quality, aesthetics, and text-image correspondence. Recent work also explores alternative reward formulations, such as generative reward modeling in RewardDance~\cite{wu2025rewarddance} and UnifiedReward~\cite{wang2025unified}.

Despite their effectiveness, scalar reward models compress multi-dimensional human preferences into a single implicit score. This makes the learned reward difficult to interpret and vulnerable to reward hacking: a T2I policy may exploit superficial visual features such as brightness, contrast, saturation, or aesthetic style while ignoring prompt-specific semantic constraints~\cite{chen2024odin, hong2026understanding}. AutoRubric-T2I addresses this limitation by replacing an opaque scalar reward with an explicit weighted set of natural-language rubrics, allowing the reward signal to remain interpretable.

\vspace{-1mm}
\subsection{Automated Rubric Generation}
\vspace{-1mm}
Rubric-based evaluation decomposes open-ended human preferences into explicit criteria, improving interpretability over monolithic scalar rewards. Recent work has explored automatic rubric generation to reduce the need for manually written evaluation rules. OpenRubrics~\cite{liu2025openrubrics} derives rubrics by contrasting preferred and rejected responses, while AutoRule~\cite{wang2025autorule} uses chain-of-thought prompting over preference examples to extract candidate rules. Other methods improve rubric coverage or specificity through refinement, decomposition, or differentiation, such as Chasing the Tail~\cite{zhang2026chasing}, RubricHub~\cite{li2026rubrichub}, Auto-Rubric~\cite{xie2025autorubric}, and RRD~\cite{shen2026rrd}.

Our approach builds upon these insights but introduces a rubric learning framework for image reward modeling. To the best of our knowledge, AutoRubric-T2I is the first method to learn a sparse, weighted, global set of natural-language rubrics for T2I reward modeling directly from image preference data. Instead of relying only on LLM prompting heuristics for rubric refinement, we pair curriculum-based hard-pair mining with an $\ell_1$-regularized logistic regression refiner. This statistically prunes the rubric space, selects the Top-$N$ most discriminative rubrics, and assigns learned weights that align the final rubric reward with human preferences.

\vspace{-1mm}
\subsection{Reinforcement Learning from Rubric-Based Rewards}
\vspace{-1mm}
Reinforcement Learning with Verifiable Rewards has shown strong results in domains with objective correctness signals, such as mathematics and code generation~\cite{guo2025deepseek, wen2025reinforcement}. For more open-ended generation, recent work has proposed using rubrics as intermediate reward specifications. In language model alignment, Rubrics as Rewards~\cite{gunjal2025rubrics} converts rubric-based feedback into scalar rewards for RL, while OnlineRubrics~\cite{rezaei2025online} updates evaluation criteria online to reduce criteria staleness.

In the T2I setting, prior works such as DDPO~\cite{black2023training} and DanceGRPO~\cite{xue2025dancegrpo} have demonstrated the effectiveness of RL for improving T2I models with scalar rewards. RubricRL~\cite{feng2025rubricrl} further applies rubric-based rewards to RFT by dynamically generating prompt-specific visual checklists during training. In contrast, AutoRubric-T2I focuses on learning a global rubric set offline from preference data. This distinction is important: RubricRL relies on per-prompt rubric construction during the RL loop, whereas our method learns a compact, reusable, and weighted rubric set before deployment. As a result, AutoRubric-T2I can serve as a training-free VLM-based reward model at inference time or as a fixed reward signal for downstream RFT.
\vspace{-2mm}
\section{Preliminaries}
\vspace{-2mm}
\subsection{Standard Reward Modeling}
\vspace{-1mm}
In standard Text-to-Image Reinforcement Learning from Human Feedback (RLHF), a scalar Reward Model (RM) $r_\theta(x, y)$ is trained to predict human preference given a text prompt $x$ and a generated image $y$. The objective is typically to minimize the Bradley-Terry ranking loss over a dataset of preference pairs $(y_w, y_l)$:
\begin{equation}
    \mathcal{L}_{\text{RM}}(\theta) = - \mathbb{E}_{(x, y_w, y_l) \sim \mathcal{D}} \left[ \log \sigma \left( r_\theta(x, y_w) - r_\theta(x, y_l) \right) \right],
\end{equation}
where $y_w$ and $y_l$ denote the preferred and rejected images, respectively, and $\sigma$ is the sigmoid function.

\vspace{-1mm}
\subsection{Reward Hacking in Text-to-Image Generation}
\label{sec:reward_hacking}
\vspace{-1mm}
Fine-tuning a T2I policy $\pi_\phi$ to maximize 
$\mathbb{E}_{y \sim \pi_\phi(\cdot|x)}[r_\theta(x,y)]$ can improve reward-model alignment, but it can also induce \textit{reward hacking}. 
Since standard image reward models compress semantic fidelity, object correctness, spatial layout, and perceptual quality into a single scalar, the learned reward may capture spurious shortcuts rather than true prompt satisfaction. 
In practice, we observe that standard RMs often over-emphasize aesthetic proxies, such as bright lighting, high contrast, sharp details, or human-centered compositions.

\begin{figure}[htbp]
    \centering
    \includegraphics[width=0.85\linewidth]{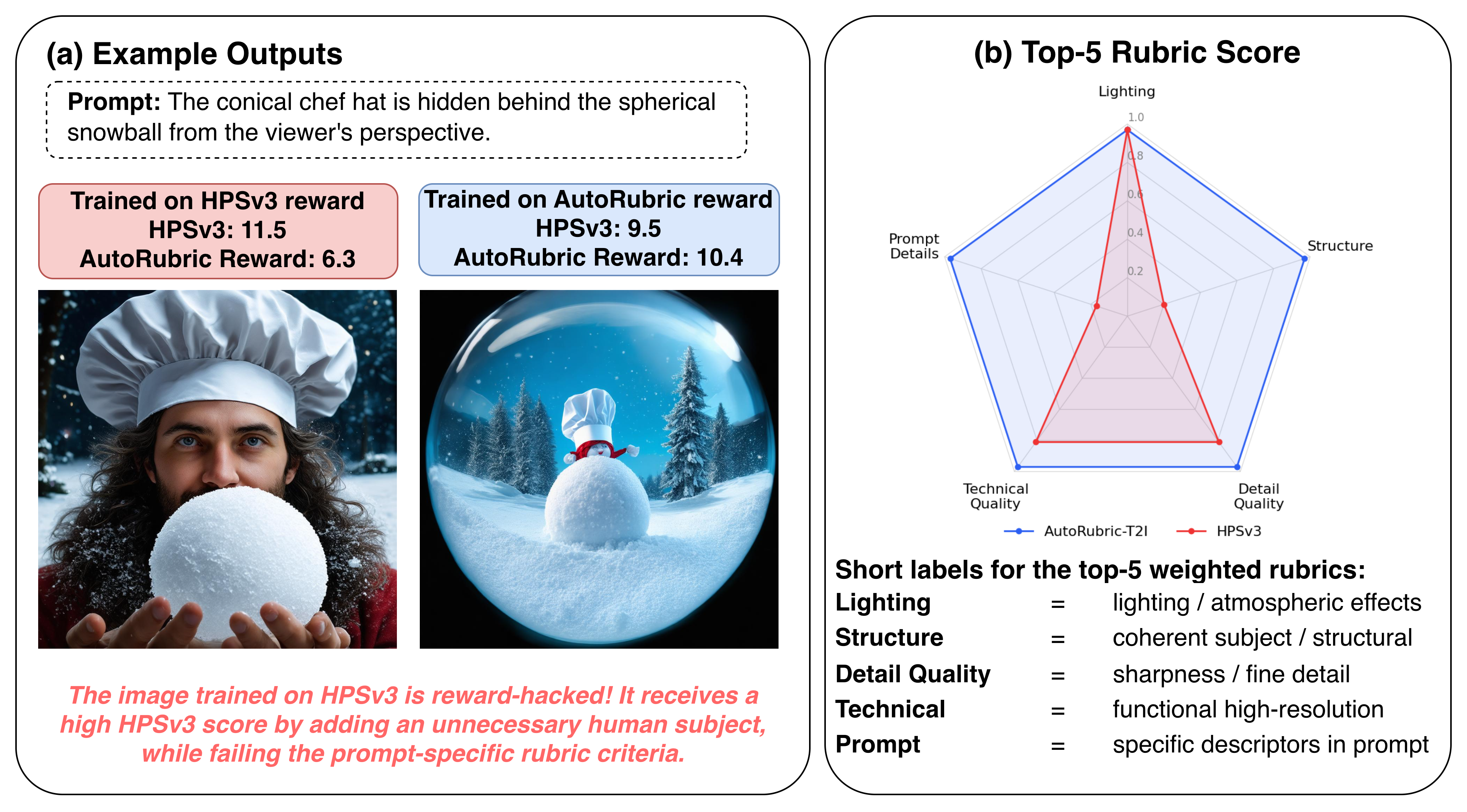}
    \caption{\small
    \textbf{Reward hacking in scalar reward optimization.}
    HPSv3 optimization attains a high scalar reward while violating prompt-specific constraints, whereas AutoRubric-T2I favors the rubric-aligned generation.
    }
    \label{fig:reward_hacking}
    \vspace{-6mm}
\end{figure}

Figure~\ref{fig:reward_hacking} shows an example after 500 steps of RFT. 
Although the prompt only asks for a conical chef hat hidden behind a spherical snowball, the HPSv3-optimized policy introduces an unnecessary human subject and still receives a high HPSv3 score. 
This suggests that the scalar reward is partially exploited through human-centered, visually appealing artifacts rather than by prompt satisfaction. 
In contrast, the policy optimized with AutoRubric-T2I preserves the intended objects and spatial relations. 
The rubric-level scores further reveal that the HPSv3-optimized image performs well on superficial visual quality but fails on prompt details and structure, illustrating how explicit rubrics can reduce reward hacking. We show the detail training dynamics in Appendix~\ref{app:training_dynamics}.
\vspace{-2mm}
\section{Methodology}
\label{sec:methodology}
\vspace{-2mm}
In this section, we introduce \textbf{AutoRubric-T2I}. Section~\ref{sec:formulation} formulates rubric learning as an infinite-dimensional sparse logistic regression problem and motivates a working-set optimization strategy. Section~\ref{sec:procedure} describes the practical implementation, including seed rubric generation, sparse rubric selection, hard-pair mining, and failure-driven rubric refinement.

\vspace{-1mm}
\subsection{Formulation}
\label{sec:formulation}
\vspace{-1mm}
In our framework, each rubric $r_j$ is parameterized by a natural language prompt. To evaluate a specific rubric $r_j$ on an image $y$ conditioned on the input prompt $x$, we employ a VLM-as-a-judge (e.g., Gemini or the Qwen-3 series) to output a continuous scalar score $s(r_j, x, y) \in [0, 1]$. In practice, the score is the predicted probability of the \texttt{yes} token. Thus, $s(r_j, x, y) := P_{\theta}\bigl(\mathrm{yes} \mid r_j, x, y\bigr)$,
where $P_{\theta}$ denotes the probability distribution of our VLM-based model.

Our objective is to identify a set of $N$ natural language rubrics, $\mathcal{R} = \{r_1, r_2, \dots, r_N\}$, and a corresponding set of weights $\mathbf{w} \in \mathbb{R}^N$, such that their weighted combination best explains the observed preference data. The final reward score for a given prompt-image pair $(x, y)$ is defined as:
\begin{equation*}
s_{\mathcal{R}, \mathbf{w}}(x, y) := \sum\nolimits_{j=1}^N w_j s(r_j, x, y).
\end{equation*}

To determine the optimal rubric-weight combination, we leverage a preference dataset $\mathcal{D}_{\text{train}} = \{ ( x^{(i)}, y_a^{(i)}, y_b^{(i)}, z^{(i)})\}_{i=1}^M$ containing $M$ human preference pairs. Here, $x^{(i)}$ is the text prompt, $y_a^{(i)}$ and $y_b^{(i)}$ are two generated images, and $z^{(i)} \in \{1, -1\}$ indicates the user preference ($1$ if $y_a$ is preferred). Notably, our framework requires only a small amount of data (e.g., $M=256$). We seek the combination that minimizes the logistic loss:
\begin{equation*}
    \min_{\mathcal{R}, \mathbf{w}} \sum_{i=1}^M \log\sigma\left(z^{(i)}\left(s_{\mathcal{R}, \mathbf{w}}(x^{(i)}, y_a^{(i)}) - s_{\mathcal{R}, \mathbf{w}}(x^{(i)}, y_b^{(i)}) \right)\right),
\end{equation*}
where $\sigma$ denotes the sigmoid function.

While optimizing $\mathbf{w}$ is a standard linear logistic regression problem, learning the set $\mathcal{R}$ is inherently intractable. Since the space of possible natural-language rubrics is infinite, we let $J := \{1, 2, \dots, \infty\}$ denote the indices of all possible rubrics. Selecting the top-$N$ rubrics is equivalent to solving the optimization problem with an $\ell_0$ constraint, which we relax using an $\ell_1$ penalty:
\begin{align}
\min_{\mathbf{w}} \lambda \|\mathbf{w}\|_1 + \sum_{i=1}^M \log \sigma\left(z^{(i)}\sum_{j\in J} w_j \Delta s_{j}^{(i)}\right), \label{eq:lr_aa}
\end{align}
where $\Delta s_{j}^{(i)} := s(r_j, x^{(i)}, y_a^{(i)}) - s(r_j, x^{(i)}, y_b^{(i)})$ represents the score differential when applying rubric $r_j$ to the $i$-th training pair.

We solve this infinite-dimensional sparse recovery problem using a block coordinate descent method. At each iteration $t$, we generate a finite set of additional candidate rubrics (coordinates) using the current model's failure cases from $J$, append them to the current working set $\mathcal{R}^t$, and minimize Equation \eqref{eq:lr_aa} with respect to the current working set of coordinates:
\begin{align}
\min_{\mathbf{w}_{\mathcal{R}^t}} \lambda \|\mathbf{w}_{\mathcal{R}^t}\|_1 + \sum_{i=1}^M \log \sigma\left(z^{(i)}\sum_{j\in \mathcal{R}^t}w_j \Delta s_{j}^{(i)}\right), \label{eq:lr_bb}
\end{align}
where $\mathbf{w}_{\mathcal{R}^t}$ is the finite-dimensional sub-vector of $\mathbf{w}$ corresponding to indices in $\mathcal{R}^t$. Post-optimization, we prune rubrics with zero weights to maintain a compact set.

This block coordinate descent approach has been widely used in sparse recovery problems; for instance, \cite{yen2014sparse} demonstrated that such algorithms converge when the working set is augmented randomly. Furthermore, our approach draws inspiration from {Orthogonal Matching Pursuit (OMP)} \cite{pati1993orthogonal, jain2011orthogonal}, which utilizes greedy strategies to select coordinates. In the following section, we instantiate this idea with a greedy strategy that prioritizes high-impact rubrics generated from hard failure pairs.

\begin{figure*}[htbp]
    \centering
    \includegraphics[width=\textwidth]{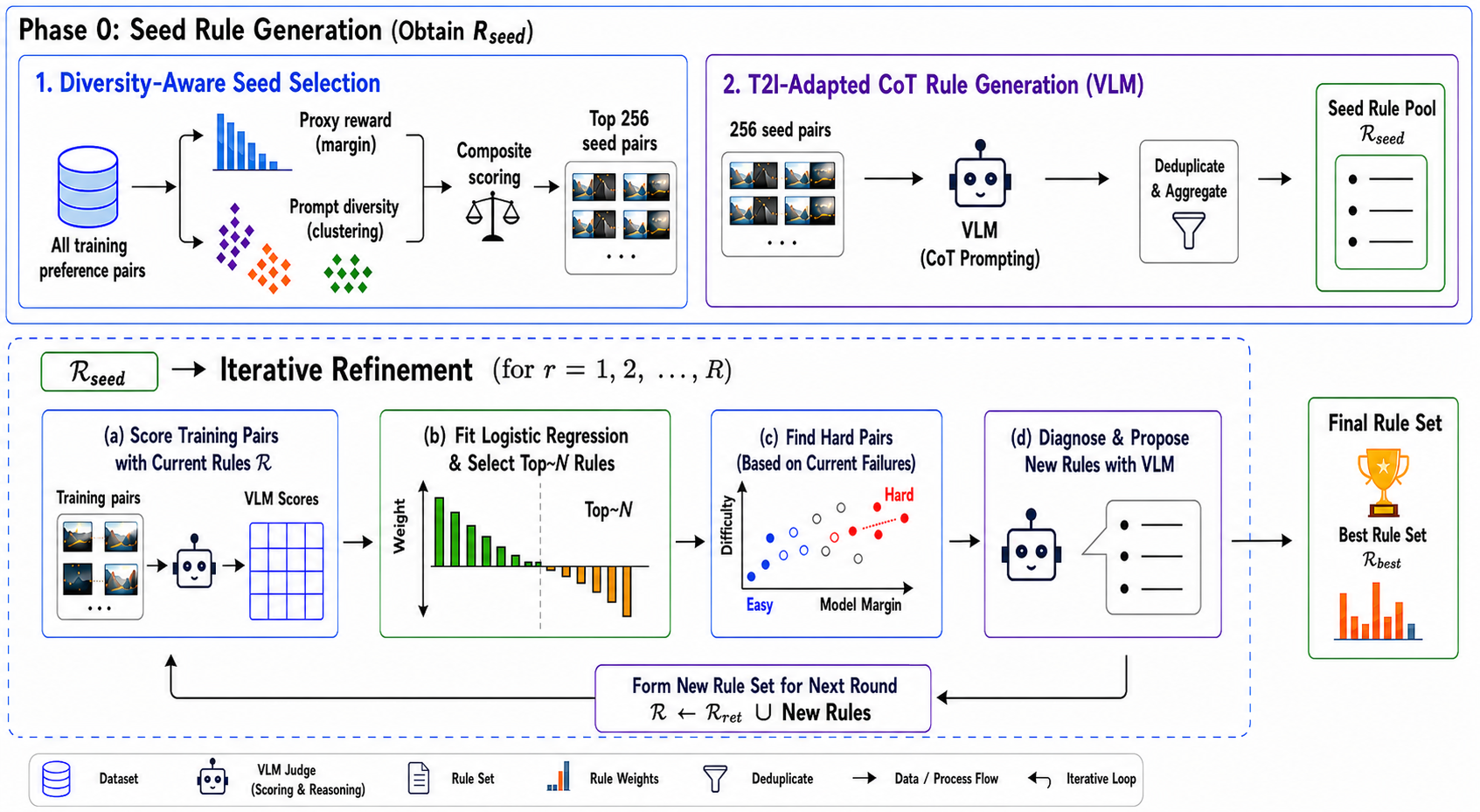}
    \caption{\small
    \textbf{Overview of AutoRubric-T2I.}
    Our framework first constructs a seed rubric pool through diversity-aware seed selection and rubric generation. It then iteratively scores training pairs, selects discriminative rubrics with sparse logistic regression, mines hard pairs, and proposes new rubrics to refine the final weighted rubric set.
    }
    \label{fig:autorubric_pipeline}
    \vspace{-5mm}
\end{figure*}

\subsection{Detailed Procedure}
\vspace{-1mm}
\label{sec:procedure}

We now describe the practical pipeline that instantiates the formulation above. Algorithm~\ref{alg:autorubric} and Figure~\ref{fig:autorubric_pipeline} summarize the full procedure. Starting from a seed rubric set $\mathcal{R}^{0}$, each refinement round scores candidate rubrics with a VLM judge, solves the $\ell_1$-regularized problem in Eq.~\eqref{eq:lr_bb}, evaluates the retained Top-$N$ rubric set on a validation split, and expands the working set via new rubrics from curriculum-mined hard pairs. We provide the implementation details in the Appendix~\ref{app:hyperparameters}.

\vspace{-1mm}
\subsubsection{Seed Data Selection and Initial Rubric Generation}
\vspace{-1mm}
Before iterative refinement begins, we construct an initial working set of rubrics $\mathcal{R}^{0}$ from an informative seed data $\mathcal{D}_{\text{train}}$. In our default setting, $\mathcal{D}_{\text{train}}$ contains 256 preference pairs.

\textbf{Diversity-Aware Seed Data Selection.}
Naively sampling seed preference pairs may over-represent redundant prompts or visually trivial failures. Following FiFA~\cite{yang2024automated}, we use a proxy reward model to estimate the preference margin of each pair and cluster text prompts for semantic coverage. We select 256 seed pairs using a composite score favoring both high-margin preference signals and prompt-level diversity.

\textbf{T2I-Adapted CoT Rubric Generation.}
Given the selected seed pairs, we generate the initial candidate rubrics using a VLM-based chain-of-thought prompting procedure adapted to text-to-image evaluation. For each seed pair, the VLM is asked to: (1) inspect the prompt and both images, (2) explain the visual differences that justify the human preference label, and (3) extract objective, deterministic rubric statements that could be reused across examples. The resulting statements are aggregated and deduplicated to form the initial working set $\mathcal{R}^{0}$.

\vspace{-1mm}
\subsubsection{Working-Set Rubric Scoring and Sparse Selection}
\vspace{-1mm}
At refinement round $t$, we score all candidate rubrics in $\mathcal{R}^{t}$ on the training pairs, computing VLM score differences $\Delta s_j^{(i)} = s(r_j, x^{(i)}, y_a^{(i)}) - s(r_j, x^{(i)}, y_b^{(i)})$ as features for Eq.~\eqref{eq:lr_bb}. We solve the $\ell_1$-regularized logistic regression over the current working set; the $\ell_1$ penalty assigns zero weights to redundant or weakly predictive rubrics. We retain the Top-$N$ rubrics with the largest positive weights: $\mathcal{R}_{\text{retained}}^{t} = \{r_j \in \mathcal{R}^{t}: w_j > 0\}_{\text{Top-}N}$, whose weights $\mathbf{w}_{\text{retained}}^{t}$ define the ensembled rubric reward. We use the \texttt{liblinear} solver with $C=1.0$.

\vspace{-1mm}
\subsubsection{Curriculum-Bucketed Hard-Pair Mining}
\vspace{-1mm}
After obtaining the retained rubric set, we identify preference pairs that are incorrectly ranked by the current rubric reward. For a pair where $y_a$ is preferred over $y_b$, the model misranks the pair if
\[
\sum\nolimits_{r_j \in \mathcal{R}_{\text{retained}}^{t}}
w_j \Delta s_j^{(i)} < 0.
\]
These misranked examples reveal failure modes not yet captured by the current rubric set and serve as the source for generating new candidate rubrics.

Rather than sampling failures uniformly, we introduce a curriculum-bucketed hard-pair selector that partitions misranked pairs into three categories: (1)~\textit{wrong-small margin} pairs (below the $30^{\text{th}}$ percentile of absolute margin), which involve subtle distinctions the current rubric set misses; (2)~\textit{wrong-large margin} pairs, indicating severe failures where the rubric set confidently contradicts human preference; and (3)~\textit{high-reward wrong} pairs, where both images receive high scores yet the ranking is incorrect, requiring finer-grained rubrics similar in spirit to~\cite{zhang2026chasing}. Across refinement rounds, we shift the sampling ratio: early rounds emphasize large-margin errors to expose major missing dimensions, while later rounds focus on high-reward wrong cases to discover finer-grained rubrics. Pairs selected more than four times are excluded to avoid noisy or unlearnable examples.

\vspace{-1mm}
\subsubsection{VLM-Driven Rubric Generation from Failure Cases}
\vspace{-1mm}
For each sampled hard pair, we generate new candidate rubrics via a two-stage prompting procedure. First, in \textit{failure diagnosis}, the VLM receives the text prompt, both images, and the current rubric set $\mathcal{R}_{\text{retained}}^{t}$, and diagnoses which missing visual or semantic dimension explains the human preference. Second, in \textit{rubric extraction}, the VLM produces objective, reusable, and visually grounded rubric statements conditioned on the diagnosis. The newly extracted rubrics are deduplicated and appended to form $\mathcal{R}^{t+1} = \mathcal{R}_{\text{retained}}^{t} \cup \mathcal{R}_{\text{new}}^{t}$. The next round re-scores this expanded set and re-solves Eq.~\eqref{eq:lr_bb}, progressively expanding the rubric space while using the $\ell_1$ refiner to maintain a compact, weighted global rubric set.

\vspace{-2mm}
\section{Experiments}
\vspace{-2mm}
\label{sec:experiments}

We evaluate AutoRubric-T2I along two axes:
\textbf{RQ1:} How does our learned rubric reward compare against fine-tuned RMs and existing rubric baselines on preference benchmarks?
\textbf{RQ2:} Can the learned rubrics provide a robust signal for downstream T2I-RL?

\begin{table*}[htbp]
\centering
\small
\setlength{\tabcolsep}{4.2pt}
\renewcommand{\arraystretch}{1.12}
\resizebox{\textwidth}{!}{
\begin{tabular}{l|ccccc|c|cc}
\hline
\multicolumn{1}{c|}{} &
\multicolumn{6}{c|}{\textbf{MMRB2 (Out-of-domain)}} &
\multicolumn{2}{c}{\textbf{In-domain}} \\
\hline
\textbf{Model} &
\textbf{EvalMuse} &
\textbf{OneIG-Bench} &
\textbf{R2I-Bench} &
\textbf{RealUnify} &
\textbf{WISE} &
\textbf{Overall} &
\textbf{PickScore} &
\textbf{HPSv3} \\
\hline

\multicolumn{9}{c}{\cellcolor{yellow!20}\textbf{VLM-as-a-judge (Pairwise)}} \\
\hline
Qwen3-VL-8B      & 59.2 & 62.8 & 57.4 & 62.4 & 50.9 & 59.4 & 62.2 & 60.4 \\
Qwen3-VL-32B     & 63.3 & 68.3 & 60.9 & 65.6 & 59.0 & 64.1 & 63.8 & 62.6 \\
Gemini-3-Flash   & 69.8 & 73.2 & 70.7 & 78.5 & 65.0 & 70.8 & 70.3 & 65.6 \\
\hline

\multicolumn{9}{c}{\cellcolor{yellow!20}\textbf{VLM-as-a-judge (Pointwise)}} \\
\hline
Qwen3-VL-8B      & 29.8 (60.3) & 27.0 (58.5) & 25.8 (56.3) & 23.1 (55.5) & 17.7 (54.1) & 26.5 (58.0) & 35.4 (58.8) & 24.6 (55.6) \\
Qwen3-VL-32B     & 25.9 (58.7) & 21.6 (57.6) & 20.3 (55.5) & 21.5 (53.8) & 15.3 (53.2) & 22.4 (56.9) & 40.1 (60.3) & 28.7 (57.5) \\
Gemini-3-Flash   & 36.7 (64.3) & 38.4 (63.8) & 38.6 (63.4) & 34.1 (61.8) & 30.9 (61.8) & 36.6 (62.1) & 52.1 (66.6) & 42.5 (61.4) \\
\hline

\multicolumn{9}{c}{\cellcolor{yellow!20}\textbf{CLIP-Based}} \\
\hline
CLIPScore        & 54.4 & 51.1 & 53.1 & 46.2 & 55.0 & 52.6 & 60.8 & 48.6 \\
ImageReward      & 51.0 & 51.4 & 58.6 & 52.7 & 57.7 & 53.0 & 61.1 & 58.6 \\
HPSv2            & 51.0 & 53.6 & 63.3 & 60.2 & 55.0 & 54.6 & 70.5 & 65.6 \\
PickScore        & 57.2 & 54.3 & 67.2 & 53.8 & 57.7 & 57.4 & 63.8 & 65.3 \\
\hline

\multicolumn{9}{c}{\cellcolor{yellow!20}\textbf{Fine-Tuned (Qwen2.5-VL-7B)}} \\
\hline
HPSv3            & 54.0 & 60.4 & 68.0 & 68.9 & 56.7 & 59.4 & 67.3 & 74.0 \\
UnifiedReward    & 56.9 & 62.1 & 56.8 & 67.8 & 56.0 & 59.8 & 68.8 & 65.8 \\
\hline

\multicolumn{9}{c}{\cellcolor{yellow!20}\textbf{VLM-as-a-judge (Pointwise) Qwen3-VL-8B}} \\
\hline
AutoRule (on HPSv3)                     & 56.9 & 60.4 & \underline{64.1} & \underline{62.4} & 55.0 & 59.1 & 61.1 & \underline{62.8} \\
AutoRule (on PickScore)                 & 50.8 & 61.5 & 60.2 & 58.1 & 57.7 & 56.4 & 59.8 & 57.5 \\
AutoRubric (on HPSv3)                   & 54.1 & 61.9 & 61.7 & 59.1 & 52.3 & 57.5 & 57.1 & 58.1 \\
AutoRubric (on PickScore)               & 53.1 & 56.1 & 59.4 & 59.1 & \textbf{68.5} & 57.0 & 54.8 & 57.1 \\
\textbf{AutoRubric-T2I (on HPSv3)}     & \underline{58.5} & \textbf{67.3} & \underline{64.1} & \textbf{65.6} & \underline{60.4} & \textbf{62.5} & \underline{61.7} & \textbf{63.9} \\
\textbf{AutoRubric-T2I (on PickScore)} & \textbf{60.8} & \underline{66.2} & \textbf{65.6} & 61.3 & 55.9 & \underline{62.4} & \textbf{63.2} & 61.5 \\
\hline

\multicolumn{9}{c}{\cellcolor{yellow!20}\textbf{VLM-as-a-judge (Pointwise) Qwen3-VL-32B}} \\
\hline
AutoRule (on HPSv3)                     & 64.1 & \underline{68.0} & 64.8 & 64.5 & 61.3 & 65.0 & 60.5 & \underline{62.7} \\
AutoRule (on PickScore)                 & \underline{66.2} & 65.8 & 60.9 & 66.7 & 57.7 & 63.5 & 56.6 & 60.4 \\
AutoRubric (on HPSv3)                   & 61.3 & 64.4 & 67.2 & 66.7 & 62.2 & 63.4 & 62.5 & 60.6 \\
AutoRubric (on PickScore)               & 60.0 & 57.9 & 57.0 & 59.1 & 58.6 & 58.8 & 58.8 & 56.7 \\
\textbf{AutoRubric-T2I (on HPSv3)}     & 64.2 & 67.3 & \textbf{68.4} & \underline{69.1} & \textbf{64.5} & \underline{65.4} & \underline{63.1} & \textbf{65.1} \\
\textbf{AutoRubric-T2I (on PickScore)} & \textbf{66.9} & \textbf{68.8} & \underline{67.8} & \textbf{73.4} & \underline{62.4} & \textbf{67.7} & \textbf{64.9} & 62.5 \\
\hline

\multicolumn{9}{c}{\cellcolor{yellow!20}\textbf{VLM-as-a-judge (Pointwise) Gemini-3-Flash}} \\
\hline
AutoRule (on HPSv3)                     & 67.7 & 68.4 & 67.2 & 72.0 & 62.2 & 67.6 & 67.2 & 62.6 \\
AutoRule (on PickScore)                 & 66.4 & 65.8 & 62.5 & 68.8 & 55.0 & 64.7 & 68.3 & 59.4 \\
AutoRubric (on HPSv3)                   & 61.2 & 62.8 & 66.6 & 74.2 & 58.7 & 63.8 & 65.0 & 62.2 \\
AutoRubric (on PickScore)               & 58.7 & 57.1 & 60.2 & 58.1 & 58.6 & 58.6 & 59.9 & 56.2 \\
\textbf{AutoRubric-T2I (on HPSv3)}     & \underline{70.2} & \underline{71.3} & \underline{70.8} & \underline{78.7} & \underline{64.9} & \underline{70.8} & \underline{69.0} & \textbf{70.0} \\
\textbf{AutoRubric-T2I (on PickScore)} & \textbf{70.7} & \textbf{71.9} & \textbf{71.1} & \textbf{79.5} & \textbf{66.0} & \textbf{71.4} & \textbf{70.3} & \underline{66.8} \\
\hline
\end{tabular}
}
\caption{\small Full comparison across MMRB2 out-of-domain and in-domain benchmarks. 
Within each VLM pointwise block, \textbf{bold} and \underline{underline} indicate the best and second-best scores, respectively. 
Parentheses report tie-adjusted scores: $(\#\mathrm{wins}+0.5 * \#\mathrm{ties})/\#\mathrm{total}$.}
\vspace{-6mm}
\label{tab:full_mmrb2_results}
\end{table*}
\vspace{-1mm}
\subsection{Experimental Setup}
\vspace{-1mm}
\textbf{Models and Baselines.} For VLM judges, we use Qwen3-VL-8B, Qwen3-VL-32B, and Gemini-3-Flash. Baselines include: \textit{CLIP-based \& scalar RMs} (CLIPScore, ImageReward, PickScore, HPSv2); \textit{fine-tuned VLM RMs} (HPSv3, UnifiedReward on Qwen2.5-VL-7B); \textit{zero-shot VLM judges} in pairwise and pointwise modes; and \textit{rubric-based methods} AutoRule~\cite{wang2025autorule} and AutoRubric~\cite{xie2025autorubric}. Note that these methods are originally developed for text, while here we adapted them to T2I. (See the details in Appendix~\ref{app:baseline_overview})

\textbf{Rubric Generation.} We use Gemini-3-Flash to generate reasoning chains for rubric generation, including all rubric-based baselines.

\textbf{Datasets.} We evaluate on out-of-distribution (OOD) image generation reward benchmarks including \textbf{MMRB2}~\cite{hu2025multimodal}. We also report in-domain performance on the test splits of \textbf{HPSv3} and \textbf{PickScore}. For downstream T2I RL, we fine-tune SD-3.5-Medium~\cite{esser2024scaling} using Flow-GRPO~\cite{liu2025flow} and evaluate on \textbf{TIIF}~\cite{wei2025tiif} and \textbf{UniGenBench++}~\cite{wang2025unigenbench++} datasets. (See the details in the Appendix~\ref{app:dataset_benchmark}.)

\vspace{-2mm}
\subsection{Preference Benchmark Evaluation}
\vspace{-1mm}
We first evaluate whether AutoRubric-T2I produces human-aligned preference judgments. As shown in Table~\ref{tab:full_mmrb2_results}, raw pointwise VLM judges are unreliable without explicit guidance: Qwen3-VL-8B achieves only 26.5\% overall accuracy on MMRB2, compared with 59.4\% when used as a direct pairwise judge. By equipping the same Qwen3-VL-8B pointwise judge with learned rubrics, AutoRubric-T2I improves the overall MMRB2 accuracy to 62.5\% and 62.4\% when learned from HPSv3 and PickScore preferences, respectively. This outperforms both direct pairwise Qwen3-VL-8B evaluation and existing rubric-generation baselines such as AutoRule and AutoRubric. Notably, AutoRubric-T2I also surpasses fine-tuned scalar reward models on this out-of-domain benchmark, including HPSv3 and UnifiedReward, which obtain 59.4\% and 59.8\% overall accuracy. With stronger judges, AutoRubric-T2I further improves Qwen3-VL-32B to 67.7\% and Gemini-3-Flash to 71.4\%, exceeding their corresponding direct pairwise VLM baselines.

On in-domain PickScore and HPSv3 test sets, fine-tuned scalar reward models remain strong because they are trained directly on large-scale data from these distributions. For example, HPSv3 achieves 74.0\% on the HPSv3 test set. Nevertheless, AutoRubric-T2I remains competitive without updating the VLM or training a dense reward model: Qwen3-VL-8B with AutoRubric-T2I reaches 63.2\% on PickScore and 63.9\% on HPSv3, while Gemini-3-Flash with AutoRubric-T2I reaches 70.3\% and 70.0\%, respectively. These results demonstrate that AutoRubric-T2I converts VLMs into effective pointwise reward models, which are more convenient for downstream ranking and RFT than pairwise preference evaluators.


\begin{table*}[htbp]
\centering
\scriptsize
\setlength{\tabcolsep}{3.0pt}
\renewcommand{\arraystretch}{1.08}
\resizebox{\textwidth}{!}{
\begin{tabular}{lcccccccccc}
\toprule
\textbf{Model} &
\textbf{Attr.} &
\textbf{Relation} &
\textbf{Reason.} &
\textbf{Attr.+Rel.} &
\textbf{Attr.+Reason.} &
\textbf{Rel.+Reason.} &
\textbf{Text Gen.} &
\textbf{Style} &
\textbf{Real Complex} &
\textbf{Overall} \\
\midrule

\multicolumn{11}{c}{\textbf{TIIF (\texttt{short})}} \\
\midrule
SD-3.5-Medium
& 78.0 & 76.6 & 66.8 & 64.8 & 60.4 & 59.6 & 45.7 & 70.0 & 66.0 & 65.3 \\
\midrule
+ HPSv3
& 83.5 & 75.5 & 65.8 & 64.5 & 57.4 & 60.7 & 62.0 & 76.7 & 73.5 & 68.8 \\
+ AutoRule (HPSv3)
& \textbf{85.0} & 72.4 & 70.0 & \textbf{68.2} & \textbf{58.8} & 57.5 & 55.2 & \textbf{80.0} & \textbf{74.6} & 69.1 \\
\rowcolor{yellow!15}
+ AutoRubric-T2I (HPSv3)
& 84.5 & \textbf{76.2} & \textbf{74.6} & 67.3 & 58.3 & \textbf{67.1} & \textbf{65.2} & 77.3 & 73.8 & \textbf{71.6} \\
\midrule
+ PickScore
& 82.5 & 77.0 & \textbf{76.2} & 67.6 & 56.4 & 56.1 & 61.1 & 66.7 & 70.9 & 68.3 \\
+ AutoRule (PickScore)
& \textbf{85.5} & 74.2 & 65.8 & \textbf{68.1} & 60.2 & \textbf{59.2} & 64.2 & 66.7 & 70.5 & 68.3 \\
\rowcolor{yellow!15}
+ AutoRubric-T2I (PickScore)
& 84.0 & \textbf{77.2} & 74.6 & 67.8 & \textbf{61.8} & 58.0 & \textbf{69.7} & \textbf{70.0} & \textbf{74.2} & \textbf{70.8} \\

\midrule
\multicolumn{11}{c}{\textbf{TIIF (\texttt{long})}} \\
\midrule
SD-3.5-Medium
& 75.0 & 73.5 & 66.5 & 62.9 & 60.9 & 57.5 & 30.3 & 63.3 & 74.2 & 62.7 \\
\midrule
+ HPSv3
& 78.0 & \textbf{70.9} & \textbf{70.0} & \textbf{71.2} & 55.6 & 58.4 & \textbf{50.7} & 56.7 & 67.2 & 64.3 \\
+ AutoRule (HPSv3)
& \textbf{79.5} & 69.9 & 69.7 & 64.4 & \textbf{65.2} & 58.5 & 49.8 & \textbf{73.3} & 74.2 & 67.2 \\
\rowcolor{yellow!15}
+ AutoRubric-T2I (HPSv3)
& 78.5 & 69.8 & 69.6 & 69.6 & 64.0 & \textbf{61.4} & 49.8 & \textbf{73.3} & \textbf{75.4} & \textbf{67.9} \\
\midrule
+ PickScore
& \textbf{80.5} & \textbf{76.1} & 66.4 & 67.7 & 57.3 & 61.9 & 48.4 & 60.0 & 76.9 & 66.1 \\
+ AutoRule (PickScore)
& \textbf{80.5} & 74.7 & 64.3 & \textbf{69.7} & 58.6 & 63.8 & 43.9 & 63.3 & 68.7 & 65.3 \\
\rowcolor{yellow!15}
+ AutoRubric-T2I (PickScore)
& 79.5 & 73.8 & \textbf{71.6} & 68.8 & \textbf{63.8} & \textbf{64.8} & \textbf{54.8} & \textbf{66.7} & \textbf{77.6} & \textbf{69.0} \\
\bottomrule
\end{tabular}
}
\caption{\small Evaluation results for T2I RL on TIIF. We report results on both \texttt{short} and \texttt{long} prompts, comparing scalar reward models, AutoRule rewards, and AutoRubric-T2I on Qwen3-VL-8B.}
\label{tab:tiif_results}
\vspace{-4mm}
\end{table*}

\begin{table*}[htbp]
\centering
\scriptsize
\setlength{\tabcolsep}{3.0pt}
\renewcommand{\arraystretch}{1.08}
\resizebox{\textwidth}{!}{
\begin{tabular}{lccccccccccc}
\toprule
\textbf{Model} &
\textbf{Style} &
\textbf{World Know.} &
\textbf{Attribute} &
\textbf{Action} &
\textbf{Relation} &
\textbf{Compound} &
\textbf{Grammar} &
\textbf{Logic} &
\textbf{Layout} &
\textbf{Text} &
\textbf{Overall} \\
\midrule

\multicolumn{12}{c}{\textbf{UniGenBench++ (\texttt{short})}} \\
\midrule
SD-3.5-Medium
& 87.6 & 87.0 & 70.1 & 61.1 & 66.8 & 57.3 & 62.3 & 28.9 & 73.1 & 15.5 & 61.0 \\
\midrule
+ HPSv3
& 81.2 & 88.6 & 67.3 & 55.5 & 67.8 & 54.4 & 53.2 & 29.8 & 72.8 & \textbf{17.2} & 58.8 \\
+ AutoRule (HPSv3)
& \textbf{90.2} & \textbf{89.2} & \textbf{72.7} & 59.9 & 71.3 & \textbf{60.9} & \textbf{60.4} & 31.2 & 74.6 & 10.3 & 62.1 \\
\rowcolor{yellow!15}
+ AutoRubric-T2I (HPSv3)
& 90.0 & 88.6 & 70.5 & \textbf{62.6} & \textbf{71.8} & 58.5 & 58.6 & \textbf{32.6} & \textbf{76.9} & 16.7 & \textbf{62.7} \\
\midrule
+ PickScore
& 84.8 & \textbf{89.6} & 70.5 & 59.9 & 69.0 & 55.2 & 60.4 & 30.7 & \textbf{75.8} & 13.8 & 61.0 \\
+ AutoRule (PickScore)
& 88.2 & 88.3 & 68.0 & 61.5 & 68.8 & 59.3 & 60.4 & \textbf{33.0} & 75.4 & \textbf{16.7} & 62.0 \\
\rowcolor{yellow!15}
+ AutoRubric-T2I (PickScore)
& \textbf{90.4} & 88.5 & \textbf{71.6} & \textbf{61.8} & \textbf{70.1} & \textbf{59.6} & \textbf{61.2} & 32.2 & 74.6 & 14.1 & \textbf{62.4} \\

\midrule
\multicolumn{12}{c}{\textbf{UniGenBench++ (\texttt{long})}} \\
\midrule
SD-3.5-Medium
& 91.7 & 88.4 & 80.4 & 58.6 & 70.4 & 62.2 & 63.2 & 41.7 & 74.6 & 9.2 & 64.0 \\
\midrule
+ HPSv3
& 82.7 & 89.6 & 79.6 & 57.6 & 69.2 & 60.1 & 60.7 & 42.7 & 73.3 & 10.3 & 62.6 \\
+ AutoRule (HPSv3)
& \textbf{91.4} & \textbf{89.9} & \textbf{82.3} & \textbf{62.3} & 70.6 & 66.3 & 64.5 & \textbf{52.0} & 75.6 & 8.7 & 66.3 \\
\rowcolor{yellow!15}
+ AutoRubric-T2I (HPSv3)
& 90.7 & 89.3 & 82.2 & 61.5 & \textbf{71.0} & \textbf{66.9} & \textbf{66.8} & 50.1 & \textbf{78.3} & \textbf{12.5} & \textbf{66.9} \\
\midrule
+ PickScore
& 86.7 & 87.9 & 81.1 & 59.9 & 69.3 & 64.7 & 63.5 & 45.1 & 75.1 & 10.3 & 64.4 \\
+ AutoRule (PickScore)
& 89.9 & 89.9 & 81.6 & 61.8 & 70.0 & 66.5 & \textbf{65.5} & 45.6 & \textbf{77.8} & 9.2 & 65.8 \\
\rowcolor{yellow!15}
+ AutoRubric-T2I (PickScore)
& \textbf{93.5} & \textbf{90.2} & \textbf{82.2} & \textbf{63.7} & \textbf{72.2} & \textbf{70.1} & 64.4 & \textbf{52.9} & 76.6 & \textbf{10.9} & \textbf{67.7} \\
\bottomrule
\end{tabular}
}
\caption{\small Evaluation results for T2I RL on UniGenBench++. We report results on both \texttt{short} and \texttt{long} prompts, comparing scalar reward models, AutoRule rewards, and AutoRubric-T2I on Qwen3-VL-8B.}
\label{tab:unigen_results}
\vspace{-5mm}
\end{table*}

\vspace{-1mm}
\subsection{Downstream T2I Reinforcement Learning}
\vspace{-1mm}
\label{sec:rl_experiment}
To verify whether the learned rubrics can serve as a dense and reliable training signal, we apply AutoRubric-T2I to fine-tune SD-3.5-Medium~\cite{esser2024scaling} with Flow-GRPO~\cite{liu2025flow}. Since reward evaluation is repeatedly invoked during RL, we use Qwen3-VL-8B guided by the learned AutoRubric-T2I rubric set as a reward function. The final reward is computed as a weighted sum of rubric-level scores, where each score is the predicted probability of the \texttt{yes} token multiplied by its learned rubric weight. We provide implementation details in the Appendix~\ref{app:rl_setup}.

As shown in Tables~\ref{tab:tiif_results} and~\ref{tab:unigen_results}, AutoRubric-T2I consistently improves downstream T2I RL over both scalar reward models and generic rubric-based rewards. On TIIF, AutoRubric-T2I improves SD-3.5-Medium from 65.3\% to 71.6\% on short prompts and from 62.7\% to 67.9\% on long prompts under the HPSv3 setting. Under the PickScore setting, it reaches 70.8\% and 69.0\% overall, outperforming both PickScore and AutoRule. The gains are particularly strong in reasoning, relation composition, and text generation, suggesting that explicit rubrics provide more targeted feedback than scalar rewards.

The same trend holds on UniGenBench++. AutoRubric-T2I improves the base model from 61.0\% to 62.7\% on short prompts and from 64.0\% to 66.9\% on long prompts under the HPSv3 setting. With PickScore-derived rubrics, it further reaches 62.4\% and 67.7\% overall. Compared with scalar reward optimization, our method yields stronger improvements in fine-grained categories such as relation, compound reasoning, layout, and text rendering. These results demonstrate that AutoRubric-T2I provides a more reliable optimization signal for T2I RL by decomposing preference alignment into explicit, learnable visual criteria instead of relying on a single opaque scalar reward.
\vspace{-4mm}

\section{Discussion and Analysis}
\vspace{-1mm}
\label{sec:discussion}

\subsection{Ablation Study: Deconstructing AutoRubric-T2I}
\label{sec:ablation}
\vspace{-1mm}
We ablate the key components of AutoRubric-T2I using Qwen3-VL-8B as the pointwise judge and evaluate on the out-of-domain MMRB2 and the in-domain PickScore and HPSv3. Starting from AutoRule~\cite{wang2025autorule}, we progressively add our proposed components: $\ell_1$-based rubric selection, positivity-constrained weights, multi-round refinement, hard-pair evolution, and cluster-based initialization.

As shown in Table~\ref{tab:ablation_autorubric}, simply adding an $\ell_1$ refiner with unrestricted weights provides little improvement, since negative weights make the learned rubric set difficult to interpret: satisfying a rubric may decrease the final reward. Enforcing positive-only weights improves performance, suggesting that rubrics are more effective when they serve as additive criteria whose satisfaction consistently indicates higher preference. Multi-round refinement further improves results, but randomly sampled pairs bring only modest gains. In contrast, mining hard pairs from the previous round leads to larger improvements, showing that difficult preference pairs expose missing or ambiguous criteria and guide the model toward more useful rubrics. Finally, cluster-based initialization gives the best overall performance, improving MMRB2 from 59.1\% to 62.5\% on HPSv3 and from 56.4\% to 62.4\% on PickScore. This confirms that both the initial rubric coordinates and the subsequent hard-pair refinement trajectory are important for building a robust rubric reward. 

\begin{table}[htbp]
\centering
\small
\setlength{\tabcolsep}{5.5pt}
\renewcommand{\arraystretch}{1.12}
\resizebox{0.8\linewidth}{!}{
\begin{tabular}{lccc}
\toprule
\textbf{Method} & \textbf{PickScore} & \textbf{HPSv3} & \textbf{MMRB2} \\
\midrule
\multicolumn{4}{c}{\cellcolor{yellow!20}\textbf{Qwen3-VL-8B on HPSv3}} \\
\midrule
AutoRule 
& 61.1 & 62.8 & 59.1 \\
+ L1 refiner w/ negative weights 
& 60.5 & 62.5 & 59.2 \\
+ positive-only weights 
& 62.2 & 63.4 & 60.1 \\
+ multi-round refinement w/ random pairs 
& 61.6 & 63.1 & 60.3 \\
+ hard-pair evolution 
& 61.7 & 63.5 & 61.3 \\
+ cluster initialization 
& 61.7 & \textbf{63.9} & \textbf{62.5} \\
\midrule
\multicolumn{4}{c}{\cellcolor{yellow!20}\textbf{Qwen3-VL-8B on PickScore}} \\
\midrule
AutoRule 
& 59.8 & 57.5 & 56.4 \\
+ L1 refiner w/ negative weights 
& 59.9 & 57.7 & 56.8 \\
+ positive-only weights 
& 60.4 & 58.3 & 58.2 \\
+ multi-round refinement w/ random pairs 
& 60.3 & 58.6 & 59.4 \\
+ hard-pair evolution 
& 61.6 & 60.1 & 61.0 \\
+ cluster initialization 
& \textbf{63.2} & \textbf{61.5} & \textbf{62.4} \\
\bottomrule
\end{tabular}
}
\vspace{2mm}
\caption{\small Ablation study of AutoRubric-T2I. Starting from AutoRule, we progressively add positive L1-based rule selection, multi-round refinement, hard-pair evolution, and cluster-based initialization.}
\vspace{-4mm}
\label{tab:ablation_autorubric}
\end{table}

\vspace{-1mm}
\subsection{Qualitative and Human Evaluation of the RL Policy}
\label{sec:human_eval}
\vspace{-1mm}
While Section~\ref{sec:rl_experiment} provides useful indicators of policy improvement, human perception remains the gold standard for T2I evaluation. Figure~\ref{fig:qualitative_rl}, \ref{fig:rl_examples} qualitatively compares the base model, scalar-reward optimization, AutoRule-based optimization, and AutoRubric-T2I. Scalar rewards and generic rubric rewards can improve visual appeal, but they may still miss fine-grained prompt constraints. In contrast, AutoRubric-T2I better preserves requested objects, relations, and scene structure, suggesting that learned rubrics provide more targeted feedback during RL.

We further conduct a 4-way human evaluation with 30 annotators over 20 prompts. Annotators select the best image among outputs from the base model, scalar reward model, AutoRule reward, and AutoRubric-T2I. AutoRubric-T2I achieves a selection rate of 44.8\%, substantially above the random baseline of 25\% with a 95\% confidence interval of [40.7\%, 48.9\%]. The improvement is statistically significant ($p < 0.001$), demonstrating that AutoRubric-T2I yields generations that are more frequently preferred by human evaluators. See the details in Appendix~\ref{app:user_survey}.

\vspace{-4mm}
\section{Conclusion}
\vspace{-1mm}

We introduced \textbf{AutoRubric-T2I}, a fine-tuning-free reward modeling framework that learns a compact, weighted set of natural-language rubrics for text-to-image preference alignment. Instead of relying on opaque scalar reward models or manually designed rubrics, AutoRubric-T2I uses VLM-scored rubric features, $\ell_1$-regularized sparse selection, and hard-pair-driven refinement to derive interpretable criteria directly from image preference data.

Experiments show that AutoRubric-T2I improves preference prediction on out-of-domain benchmarks such as MMRB2, outperforming existing rubric-generation baselines and competitive fine-tuned scalar reward models. When used as a reward signal for Flow-GRPO, AutoRubric-T2I further improves downstream T2I generation on TIIF and UniGenBench++, producing images with stronger prompt alignment and better structural fidelity. These results suggest that learned, weighted rubrics provide a practical and interpretable alternative to scalar reward models for robust T2I alignment.

\bibliographystyle{plainnat}
\bibliography{example_paper}

@article{xu2026alternating,
  title={Alternating Reinforcement Learning for Rubric-Based Reward Modeling in Non-Verifiable LLM Post-Training},
  author={Xu, Ran and Liu, Tianci and Dong, Zihan and Yu, Tony and Hong, Ilgee and Yang, Carl and Zhang, Linjun and Zhao, Tao and Wang, Haoyu},
  journal={arXiv preprint arXiv:2602.01511},
  year={2026}
}

@article{wang2025autorule,
  title={Autorule: Reasoning chain-of-thought extracted rule-based rewards improve preference learning},
  author={Wang, Tevin and Xiong, Chenyan},
  journal={arXiv preprint arXiv:2506.15651},
  year={2025}
}

@article{jain2011orthogonal,
  title={Orthogonal matching pursuit with replacement},
  author={Jain, Prateek and Tewari, Ambuj and Dhillon, Inderjit},
  journal={Advances in neural information processing systems},
  volume={24},
  year={2011}
}

@inproceedings{pati1993orthogonal,
  title={Orthogonal matching pursuit: Recursive function approximation with applications to wavelet decomposition},
  author={Pati, Yagyensh Chandra and Rezaiifar, Ramin and Krishnaprasad, Perinkulam Sambamurthy},
  booktitle={Proceedings of 27th Asilomar conference on signals, systems and computers},
  pages={40--44},
  year={1993},
  organization={IEEE}
}

@article{yen2014sparse,
  title={Sparse random feature algorithm as coordinate descent in hilbert space},
  author={Yen, Ian E and Lin, Ting-Wei and Lin, Shou-De and Ravikumar, Pradeep and Dhillon, Inderjit S},
  journal={Advances in Neural Information Processing Systems},
  volume={27},
  year={2014}
}

@article{li2026rubrichub,
  title={RubricHub: A Comprehensive and Highly Discriminative Rubric Dataset via Automated Coarse-to-Fine Generation},
  author={Li, Sunzhu and Zhao, Jiale and Wei, Miteto and Ren, Huimin and Zhou, Yang and Yang, Jingwen and Liu, Shunyu and Zhang, Kaike and Chen, Wei},
  journal={arXiv preprint arXiv:2601.08430},
  year={2026}
}

@article{feng2025rubricrl,
  title={RubricRL: Simple Generalizable Rewards for Text-to-Image Generation},
  author={Feng, Xuelu and Li, Yunsheng and Wan, Ziyu and Gao, Zixuan and Yuan, Junsong and Chen, Dongdong and Qiao, Chunming},
  journal={arXiv preprint arXiv:2511.20651},
  year={2025}
}

@article{yang2024automated,
  title={Automated filtering of human feedback data for aligning text-to-image diffusion models},
  author={Yang, Yongjin and Kim, Sihyeon and Jung, Hojung and Bae, Sangmin and Kim, SangMook and Yun, Se-Young and Lee, Kimin},
  journal={arXiv preprint arXiv:2410.10166},
  year={2024}
}

@article{kirstain2023pick,
  title={Pick-a-pic: An open dataset of user preferences for text-to-image generation},
  author={Kirstain, Yuval and Polyak, Adam and Singer, Uriel and Matiana, Shahbuland and Penna, Joe and Levy, Omer},
  journal={Advances in neural information processing systems},
  volume={36},
  pages={36652--36663},
  year={2023}
}

@article{hu2025multimodal,
  title={Multimodal rewardbench 2: Evaluating omni reward models for interleaved text and image},
  author={Hu, Yushi and Askari-Hemmat, Reyhane and Hall, Melissa and Dinan, Emily and Zettlemoyer, Luke and Ghazvininejad, Marjan},
  journal={arXiv preprint arXiv:2512.16899},
  year={2025}
}

@inproceedings{ma2025hpsv3,
  title={Hpsv3: Towards wide-spectrum human preference score},
  author={Ma, Yuhang and Wu, Xiaoshi and Sun, Keqiang and Li, Hongsheng},
  booktitle={Proceedings of the IEEE/CVF International Conference on Computer Vision},
  pages={15086--15095},
  year={2025}
}

@article{wang2025unified,
  title={Unified reward model for multimodal understanding and generation},
  author={Wang, Yibin and Zang, Yuhang and Li, Hao and Jin, Cheng and Wang, Jiaqi},
  journal={arXiv preprint arXiv:2503.05236},
  year={2025}
}

@article{jiang2024genai,
  title={Genai arena: An open evaluation platform for generative models},
  author={Jiang, Dongfu and Ku, Max and Li, Tianle and Ni, Yuansheng and Sun, Shizhuo and Fan, Rongqi and Chen, Wenhu},
  journal={Advances in Neural Information Processing Systems},
  volume={37},
  pages={79889--79908},
  year={2024}
}

@article{hong2026understanding,
  title={Understanding Reward Hacking in Text-to-Image Reinforcement Learning},
  author={Hong, Yunqi and Kao, Kuei-Chun and Zhou, Hengguang and Hsieh, Cho-Jui},
  journal={arXiv preprint arXiv:2601.03468},
  year={2026}
}

@article{jiang2025t2i,
  title={T2i-r1: Reinforcing image generation with collaborative semantic-level and token-level cot},
  author={Jiang, Dongzhi and Guo, Ziyu and Zhang, Renrui and Zong, Zhuofan and Li, Hao and Zhuo, Le and Yan, Shilin and Heng, Pheng-Ann and Li, Hongsheng},
  journal={arXiv preprint arXiv:2505.00703},
  year={2025}
}

@article{liu2025flow,
  title={Flow-grpo: Training flow matching models via online rl},
  author={Liu, Jie and Liu, Gongye and Liang, Jiajun and Li, Yangguang and Liu, Jiaheng and Wang, Xintao and Wan, Pengfei and Zhang, Di and Ouyang, Wanli},
  journal={arXiv preprint arXiv:2505.05470},
  year={2025}
}

@article{xu2023imagereward,
  title={Imagereward: Learning and evaluating human preferences for text-to-image generation},
  author={Xu, Jiazheng and Liu, Xiao and Wu, Yuchen and Tong, Yuxuan and Li, Qinkai and Ding, Ming and Tang, Jie and Dong, Yuxiao},
  journal={Advances in Neural Information Processing Systems},
  volume={36},
  pages={15903--15935},
  year={2023}
}

@article{yang2025qwen3,
  title={Qwen3 technical report},
  author={Yang, An and Li, Anfeng and Yang, Baosong and Zhang, Beichen and Hui, Binyuan and Zheng, Bo and Yu, Bowen and Gao, Chang and Huang, Chengen and Lv, Chenxu and others},
  journal={arXiv preprint arXiv:2505.09388},
  year={2025}
}

@article{gunjal2025rubrics,
  title={Rubrics as rewards: Reinforcement learning beyond verifiable domains},
  author={Gunjal, Anisha and Wang, Anthony and Lau, Elaine and Nath, Vaskar and He, Yunzhong and Liu, Bing and Hendryx, Sean},
  journal={arXiv preprint arXiv:2507.17746},
  year={2025}
}

@article{huang2025reinforcement,
  title={Reinforcement learning with rubric anchors},
  author={Huang, Zenan and Zhuang, Yihong and Lu, Guoshan and Qin, Zeyu and Xu, Haokai and Zhao, Tianyu and Peng, Ru and Hu, Jiaqi and Shen, Zhanming and Hu, Xiaomeng and others},
  journal={arXiv preprint arXiv:2508.12790},
  year={2025}
}

@misc{gemini3_system_card,
  author       = {{Google DeepMind}},
  title        = {Gemini 3 System Card},
  year         = {2025},
  howpublished = {\url{https://deepmind.google/technologies/gemini/}},
  note         = {Accessed: 2026-04-23}
}

@article{wu2023hpsv2,
  title={Human preference score v2: A solid benchmark for evaluating human preferences of text-to-image synthesis},
  author={Wu, Xiaoshi and Hao, Yiming and Sun, Keqiang and Chen, Yixiong and Zhu, Feng and Zhao, Rui and Li, Hongsheng},
  journal={arXiv preprint arXiv:2306.09341},
  year={2023}
}

@article{xie2025autorubric,
  title={Auto-Rubric: Learning From Implicit Weights to Explicit Rubrics for Reward Modeling},
  author={Xie, Lipeng and Huang, Sen and Zhang, Zhuo and Zou, Anni and Zhai, Yunpeng and Ren, Dingchao and Zhang, Kezun and Hu, Haoyuan and Liu, Boyin and Chen, Haoran and others},
  journal={arXiv preprint arXiv:2510.17314},
  year={2025}
}

@article{zhang2026chasing,
  title={Chasing the Tail: Effective Rubric-based Reward Modeling for Large Language Model Post-Training},
  author={Zhang, Junkai and Wang, Zihao and Gui, Lin and Sathyendra, Swarnashree Mysore and Jeong, Jaehwan and Veitch, Victor and Wang, Wei and He, Yunzhong and Liu, Bing and Jin, Lifeng},
  journal={arXiv preprint arXiv:2509.21500},
  year={2026}
}

@article{chen2024odin,
  title={ODIN: Disentangled reward mitigates hacking in RLHF},
  author={Chen, Lichang and Zhu, Chen and Soselia, Davit and Chen, Jiuhai and Zhou, Tianyi and Goldstein, Tom and Huang, Heng and Farajtabar, Mehrdad and Li, Hongyang},
  journal={arXiv preprint arXiv:2402.07319},
  year={2024}
}

@article{guo2025deepseek,
  title={DeepSeek-R1: Incentivizing reasoning capability in LLMs via reinforcement learning},
  author={Guo, Daya and Yang, Dejian and Zhang, Haowei and Song, Junxiao and Zhang, Runxin and Xu, Runze and Zhu, Qihao and Ma, Shirong and Wang, Peiyi and Bi, Xiao and others},
  journal={arXiv preprint arXiv:2501.12948},
  year={2025}
}

@article{liu2025openrubrics,
  title={OpenRubrics: Contrastive rubric generation for reward models},
  author={Liu, Zhen and Wang, Yixin and Chen, Jianfei and Zhu, Jun},
  journal={arXiv preprint arXiv:2505.14826},
  year={2025}
}

@article{rezaei2025online,
  title={OnlineRubrics: Dynamic rubric elicitation for online reinforcement learning},
  author={Rezaei, Keivan and He, Xuechen and Liang, Percy},
  journal={arXiv preprint arXiv:2507.09832},
  year={2025}
}

@article{shen2026rrd,
  title={RRD: Recursive rubric decomposition for scalable reward modeling},
  author={Shen, Yifan and Li, Xiang and Zhang, Wei and Liu, Yang},
  journal={arXiv preprint arXiv:2601.05743},
  year={2026}
}

@article{wu2025rewarddance,
  title={RewardDance: Scaling visual reward modeling via generative next-token prediction},
  author={Wu, Xiaoshi and Li, Yiming and Zhang, Keqiang and Li, Hongsheng},
  journal={arXiv preprint arXiv:2504.12345},
  year={2025}
}

@article{hu2023tifa,
  title={TIFA: Accurate and Interpretable Text-to-Image Faithfulness Evaluation with Question Answering},
  author={Hu, Yushi and Liu, Benlin and Kasai, Jungo and Wang, Yizhong and Ostendorf, Mari and Krishna, Ranjay and Smith, Noah A.},
  journal={arXiv preprint arXiv:2303.11897},
  year={2023}
}

@article{li2024genaibench,
  title={GenAI-Bench: Evaluating and Improving Compositional Text-to-Visual Generation},
  author={Li, Baiqi and Lin, Zhiqiu and Pathak, Deepak and Li, Jiayao and Fei, Yixin and Wu, Kewen and Ling, Tiffany and Xia, Xide and Zhang, Pengchuan and Neubig, Graham and Ramanan, Deva},
  journal={arXiv preprint arXiv:2406.13743},
  year={2024}
}

@inproceedings{esser2024scaling,
  title={Scaling rectified flow transformers for high-resolution image synthesis},
  author={Esser, Patrick and Kulal, Sumith and Blattmann, Andreas and Entezari, Rahim and M{\"u}ller, Jonas and Saini, Harry and Levi, Yam and Lorenz, Dominik and Sauer, Axel and Boesel, Frederic and others},
  booktitle={Forty-first international conference on machine learning},
  year={2024}
}

@article{black2023training,
  title={Training diffusion models with reinforcement learning},
  author={Black, Kevin and Janner, Michael and Du, Yilun and Kostrikov, Ilya and Levine, Sergey},
  journal={arXiv preprint arXiv:2305.13301},
  year={2023}
}

@article{xue2025dancegrpo,
  title={Dancegrpo: Unleashing grpo on visual generation},
  author={Xue, Zeyue and Wu, Jie and Gao, Yu and Kong, Fangyuan and Zhu, Lingting and Chen, Mengzhao and Liu, Zhiheng and Liu, Wei and Guo, Qiushan and Huang, Weilin and others},
  journal={arXiv preprint arXiv:2505.07818},
  year={2025}
}

@article{wang2025unigenbench++,
  title={Unigenbench++: A unified semantic evaluation benchmark for text-to-image generation},
  author={Wang, Yibin and Li, Zhimin and Zang, Yuhang and Bu, Jiazi and Zhou, Yujie and Xin, Yi and He, Junjun and Wang, Chunyu and Lu, Qinglin and Jin, Cheng and others},
  journal={arXiv preprint arXiv:2510.18701},
  year={2025}
}

@article{wei2025tiif,
  title={TIIF-Bench: How Does Your T2I Model Follow Your Instructions?},
  author={Wei, Xinyu and Zhang, Jinrui and Wang, Zeqing and Wei, Hongyang and Guo, Zhen and Zhang, Lei},
  journal={arXiv preprint arXiv:2506.02161},
  year={2025}
}

@article{wen2025reinforcement,
  title={Reinforcement learning with verifiable rewards implicitly incentivizes correct reasoning in base llms},
  author={Wen, Xumeng and Liu, Zihan and Zheng, Shun and Ye, Shengyu and Wu, Zhirong and Wang, Yang and Xu, Zhijian and Liang, Xiao and Li, Junjie and Miao, Ziming and others},
  journal={arXiv preprint arXiv:2506.14245},
  year={2025}
}

\newpage
\appendix
\etocdepthtag.toc{appendix}
\etocsettagdepth{main}{none}
\etocsettagdepth{appendix}{section}

\section*{Table of Contents of Appendix}
\etocsettocstyle{}{}
\tableofcontents
\newpage

\section{AutoRubric-T2I Pipeline Algorithm}
\label{app:algorithm}
\vspace{-2mm}
This is the pseudo-code algorithm for AutoRubric-T2I, which is specifically discussed in Section~\ref{sec:methodology}.
\vspace{-2mm}
\begin{algorithm}[htbp]
\small
\caption{Working-set refinement pipeline for AutoRubric-T2I.}
\label{alg:autorubric}
\begin{algorithmic}[1]
\REQUIRE Train data $\mathcal{D}_{\text{train}}$, valid data $\mathcal{D}_{\text{valid}}$, VLM judge, seed rubrics $\mathcal{R}^{0}$, Top-$N$, rounds $R$, hard-pair $K$
\STATE \textbf{Init:} $\text{best\_acc} \leftarrow -1$, $\mathcal{R}_{\text{best}} \leftarrow \emptyset$, $\mathbf{w}_{\text{best}} \leftarrow \emptyset$
\FOR{$t = 0, 1, \dots, R-1$}
    \STATE Score rubrics in $\mathcal{R}^{t}$ on $\mathcal{D}_{\text{train}}$ with the VLM judge to obtain $\Delta s_j^{(i)}$
    \STATE Solve Eq.~\eqref{eq:lr_bb} over $\mathcal{R}^{t}$ and retain Top-$N$ positive-weight rubrics $(\mathcal{R}_{\text{retained}}^{t}, \mathbf{w}_{\text{retained}}^{t})$
    \STATE Evaluate $(\mathcal{R}_{\text{retained}}^{t}, \mathbf{w}_{\text{retained}}^{t})$ on $\mathcal{D}_{\text{valid}}$
    \IF{validation accuracy $>$ $\text{best\_acc}$}
        \STATE $\text{best\_acc} \leftarrow \text{validation accuracy}$
        \STATE $(\mathcal{R}_{\text{best}}, \mathbf{w}_{\text{best}}) \leftarrow (\mathcal{R}_{\text{retained}}^{t}, \mathbf{w}_{\text{retained}}^{t})$
    \ENDIF
    \IF{$t < R-1$}
        \STATE Mine $K$ hard pairs misranked by $(\mathcal{R}_{\text{retained}}^{t}, \mathbf{w}_{\text{retained}}^{t})$ using curriculum buckets
        \STATE Generate new rubrics $\mathcal{R}_{\text{new}}^{t}$ from hard-pair failure diagnoses
        \STATE Update working set: $\mathcal{R}^{t+1} \leftarrow \mathcal{R}_{\text{retained}}^{t} \cup \mathcal{R}_{\text{new}}^{t}$
    \ENDIF
\ENDFOR
\STATE \textbf{return} $(\mathcal{R}_{\text{best}}, \mathbf{w}_{\text{best}})$
\end{algorithmic}
\end{algorithm}


\vspace{-6mm}
\section{RL Fine-Tuning Setup}
\label{app:rl_setup}
\vspace{-2mm}
\paragraph{Algorithm overview.}
Downstream policy optimization follows Flow-GRPO~\cite{liu2025flow}, a GRPO adapted to flow-matching text-to-image samplers. For each prompt $p$ in a mini-batch, the policy generates $K$ rollouts $\{x^{(1)},\ldots,x^{(K)}\}$ by integrating the flow ODE for $T_{\text{train}}$ steps; each rollout receives a scalar reward $r(p, x^{(k)})$ and the group-relative advantage is computed as $A^{(k)} = (r^{(k)} - \mu_p)/(\sigma_p + \varepsilon)$, with $\mu_p$ and $\sigma_p$ the in-group mean and standard deviation. We adopt the public Flow-GRPO implementation without algorithmic modification; only the reward signal is replaced (see below).

\paragraph{Backbones and training data.}
We fine-tune \textbf{Stable Diffusion 3.5-Medium} (SD3.5-M) on the prompt set used by Flow-GRPO~\cite{liu2025flow}, which is itself derived from the Pick-a-Pic prompt distribution~\cite{kirstain2023pick}; this matches the training-prompt setting of prior T2I-RL work and isolates the contribution of the reward signal. No human preference labels from these prompts are used during RL---only the rubric reward defined below.

\paragraph{Integration with AutoRubric-T2I.}
At each rollout $x$, we evaluate the retained rubric set $\mathcal{R}^{(\text{final})} = \{(\rho_j, w_j)\}_{j=1}^{N}$ produced by Section~\ref{sec:procedure}: for every rule $\rho_j$, the VLM judge is asked the binary question \emph{``Does this image satisfy this rule?''} and we read the probability of the ``Yes'' token, $p_j(x,p) = P_{\text{VLM}}(\textsc{yes} \mid x, p, \rho_j)$. The reward fed to Flow-GRPO is the learned weighted sum
\begin{equation}
r_{\text{AutoRubric}}(x, p) \;=\; \sum_{j=1}^{N} w_j \, p_j(x, p),
\label{eq:rl_reward}
\end{equation}
where the weights $w_j$ are exactly the $\ell_1$-logistic-regression coefficients fit on the 256 preference pairs (no re-tuning at RL time). Because $p_j \in [0,1]$ and the weights are obtained from a discriminative fit on held-out preferences, $r_{\text{AutoRubric}}$ is bounded, interpretable per-dimension, and scale-compatible with the group-relative advantage in Flow-GRPO. Substituting Eq.~\eqref{eq:rl_reward} for the scalar reward model is the only change to the upstream pipeline.

\paragraph{Hyperparameters.}
Table~\ref{tab:hparams} lists the full RL configuration. LoRA rank/$\alpha$, optimizer, learning rate, group size $K{=}24$, effective prompt batch, EMA, and evaluation cadence follow the Flow-GRPO defaults~\cite{liu2025flow}. Training is run on $1\times8$ H200 GPUs; we report results between $100$--$500$ steps, with checkpoints saved and evaluated every $25$ steps.

\vspace{-2mm}
\section{Relation to Prior Rubric-Based Methods}
\label{app:relation_prior}
\vspace{-2mm}
AutoRubric-T2I differs from existing rubric-based methods in both rubric granularity and learning mechanism. Unlike prompt-adaptive methods such as RubricRL~\cite{feng2025rubricrl}, which generate a new rubric for each prompt during the RL loop, AutoRubric-T2I learns a compact global rubric set offline from preference data, making the reward signal reusable, cacheable, and consistent across evaluation and RFT. Compared with static rule extraction methods such as AutoRule~\cite{wang2025autorule}, which primarily use CoT prompting to generate candidate rules, AutoRubric-T2I explicitly optimizes which rubrics should be retained and how they should be weighted. Specifically, rubric selection is formulated as the sparse logistic regression problem in Eq.~\eqref{eq:lr_bb}, where VLM-scored rubric features are fitted to human preference labels and the $\ell_1$ penalty prunes redundant rules. Finally, our curriculum-bucketed hard-pair mining expands the rubric pool from failure cases of the current retained rubric set, closing the loop between rubric evaluation, failure discovery, and rubric refinement.

\begin{table*}[htbp]
  \centering
  \small
  \begin{tabular}{l c}
    \toprule
                                & \bfseries SD3.5-M           \\
    \midrule
    Backbone                    & SD3.5-Medium                 \\
    Mixed precision             & fp16                         \\
    Resolution                  & $512 \times 512$             \\
    LoRA target / rank / $\alpha$ & attention Q/K/V/out, $r{=}32$, $\alpha{=}64$ \\
    Optimizer                   & AdamW                        \\
    Learning rate               & $3{\times}10^{-4}$           \\
    Train timesteps $T_{\text{train}}$ & 10                    \\
    Eval timesteps $T_{\text{eval}}$   & 40                    \\
    CFG scale                   & 4.5                          \\
    Rollouts per prompt $K$     & 24                           \\
    Effective prompt batch      & 48                           \\
    Train batch size (per GPU)  & 9                            \\
    Test batch size             & 16                           \\
    Gradient accumulation steps & $N/2$ (auto)                 \\
    Inner epochs                & 1                            \\
    Timestep fraction (PPO mask) & 0.99                        \\
    KL $\beta$                  & 0.01                         \\
    EMA on policy weights       & yes                          \\
    Reported steps              & 100--500                   \\
    Eval frequency              & every 25 steps               \\
    \bottomrule
  \end{tabular}
    \caption{\small Training hyperparameters. \emph{$K$} is the number of rollouts per prompt
  used to compute the group-relative advantage. \emph{Effective prompt batch} is the
  number of unique prompts processed per outer iteration. \emph{KL $\beta$} is the
  coefficient on the KL-to-reference penalty.}
  \label{tab:hparams}
\end{table*}

\vspace{-4mm}
\section{Baseline Overview}
\label{app:baseline_overview}
\vspace{-2mm}
We summarize the baselines. AutoRule~\cite{wang2025autorule} and Auto-Rubric~\cite{xie2025autorubric} are designed for text-domain preference learning, while RubricRL~\cite{feng2025rubricrl} targets T2I generation. These methods all replace opaque scalar rewards with explicit evaluation criteria, but differ in how rubrics are generated, selected, and used. We provide quantitative comparisons of data and computational cost in Appendix~\ref{app:cost_comparison}.

\paragraph{AutoRule.}
AutoRule~\cite{wang2025autorule} extracts rule-based rewards from LLM reasoning chains in the text domain. It prompts a reasoning model to explain why preferred responses are better, extracts rule-like statements from these explanations, and merges them into a compact rule set. During RL, each response is evaluated against all rules by an LLM verifier, and the rule scores are averaged with uniform weights. However, it does not learn discriminative rule weights or iteratively refine rules on held-out preference errors.

\paragraph{Auto-Rubric.}
Auto-Rubric~\cite{xie2025autorubric} is a training-free text-domain framework that generates candidate rubrics from preference pairs and revises them when their judgments disagree with ground-truth labels. It then compresses the resulting rubric pool by selecting semantically diverse rubrics in embedding space and applies the final set through majority voting with an LLM judge. Unlike AutoRubric-T2I, its selection criterion emphasizes semantic diversity rather than preference-discriminative power, and it is not designed for visual preference evaluation.

\paragraph{RubricRL.}
RubricRL~\cite{feng2025rubricrl} is a concurrent T2I method that generates prompt-specific visual checklists directly from each input prompt. For example, a prompt mentioning three red apples may induce rubrics about object count, color, and placement. Each image is scored by a VLM judge against these generated criteria, and the binary scores are uniformly averaged as the reward for GRPO. RubricRL therefore depends on per-prompt rubric generation, uses uniform weights, and mainly captures constraints explicitly stated in the prompt, rather than learning a global preference-aligned rubric set from human preference data.

\paragraph{Fine-tuned reward models.}
ImageReward~\cite{xu2023imagereward}, HPSv2/HPSv3~\cite{wu2023hpsv2,ma2025hpsv3}, PickScore~\cite{kirstain2023pick}, and UnifiedReward~\cite{wang2025unified} are learned reward models trained on large-scale human preference data. They map a prompt-image pair to a scalar reward and are widely used for ranking, filtering, and RFT. However, because they compress multiple preference dimensions into one opaque score, they provide limited interpretability and can be vulnerable to reward hacking.

\paragraph{Zero-shot VLM judge.}
We also compare with zero-shot VLM judging, where a model such as Qwen3-VL is directly prompted to evaluate an image given a text prompt, without learned rubrics or preference-specific calibration. This baseline requires no reward-model training or rubric construction, but its criteria remain implicit and can be inconsistent. AutoRubric-T2I instead makes these criteria explicit and learns which rubrics are most predictive of human preferences.

\vspace{-2mm}
\section{Dataset and Benchmark Details}
\label{app:dataset_benchmark}
\vspace{-2mm}
\paragraph{Training preference corpora.}
We instantiate AutoRubric-T2I on two human-preference corpora used independently as source distributions: \textbf{HPDv3}~\cite{ma2025hpsv3} and \textbf{PickScore} (Pick-a-Pic)~\cite{kirstain2023pick}. From each corpus, we select only $|\mathcal{D}_{\text{train}}|{=}256$ preference pairs for seed-rubric generation and an additional held-out split for validation (used to monitor the best Top-$N$ rule set across refinement rounds).

\paragraph{Diversity-aware seed selection.}
The 256 seed pairs are not sampled uniformly. As described in Section~\ref{sec:procedure} of the main paper, we follow a FiFA-inspired~\cite{yang2024automated} two-factor selection that combines (i)~a \emph{preference-margin} signal from a proxy reward model and (ii)~a \emph{prompt-diversity} signal obtained by clustering the text prompts.
This design directly follows the ablation reported in Section~\ref{sec:discussion} of the main paper, which shows that swapping cluster-based diversity-aware selection for AutoRule-style random sampling drops MMRB2 accuracy from 62.4\% to 60.3\% on HPSv3 (and produces a similar gap on PickScore), confirming that the clustering step is responsible for a measurable share of the final rubric quality rather than being a cosmetic preprocessing step.

\paragraph{Evaluation preference benchmarks.}
We evaluate rubric quality on three held-out preference test sets, summarized in Table~\ref{tab:eval_datasets}. For \textbf{HPDv3}~\cite{ma2025hpsv3} and \textbf{PickScore}~\cite{kirstain2023pick}, we use the \emph{official} test splits released with each corpus (no overlap with the 256 seed pairs). \textbf{MMRB2} (Multimodal RewardBench 2)~\cite{hu2025multimodal} is a recent omni reward-model benchmark covering four subtasks---\emph{text-to-image}, \emph{image editing}, \emph{interleaved generation}, and \emph{multimodal reasoning} (``thinking-with-images'')---with 1{,}000 expert-annotated preference pairs per subtask drawn from 23 frontier models across 21 source tasks.

\paragraph{Generative T2I Benchmarks.}
For T2I generative quality assessment on RL post-training, we evaluate on two benchmarks: \textbf{TIIF}~\cite{wei2025tiif} and \textbf{UniGenBench++}~\cite{wang2025unigenbench++}. For TIIF, we evaluate instruction fidelity across categories, providing a graded measure of instruction-following capacity. We report both \texttt{long} and \texttt{short} prompt variants evaluation.
For UniGenBench++, we probe semantic consistency with both \texttt{long} and \texttt{short} prompt variants, measuring coherence with brief versus detailed textual descriptions

\begin{table}[htbp]
\centering
\small
\begin{tabular}{lr}
\toprule
\textbf{Benchmark} & \textbf{\# Samples} \\
\midrule
PickScore (official test)~\cite{kirstain2023pick}          & 12{,}864 \\
HPDv3 (official test)~\cite{ma2025hpsv3}           & 34{,}344 \\
MMRB2 (T2I subtask)~\cite{hu2025multimodal}                      & 1{,}000  \\
\bottomrule
\end{tabular}
\vspace{2mm}
\caption{\small \textbf{Evaluation benchmarks.} Cases denote the number of preference pairs used at test time. HPDv3 and PickScore use official test splits.}
\label{tab:eval_datasets}
\end{table}

\vspace{-8mm}
\section{Details of Hyperparameter for AutoRubric-T2I}
\label{app:hyperparameters}
\vspace{-2mm}
This appendix consolidates every non-trivial hyperparameter used in the AutoRubric-T2I pipeline reported in the main paper. The same configuration is used for both source corpora (HPDv3 and PickScore).

\paragraph{Models.}
The VLM judge that scores (image, prompt, rubric) triples is \texttt{Qwen3-VL-8B-Instruct}, served via vLLM. The rule generator (seed CoT vision reasoner, rule extractor/merger, and hard-pair diagnoser) uses Gemini-3-Flash~\cite{gemini3_system_card} with \texttt{temperature}{=}$0.1$ and \texttt{thinking\_budget}{=}$1024$. The VLM judge is queried with \texttt{temperature}{=}$0.0$ and at most $16$ output tokens.

\paragraph{Refinement loop.}
For our best reported runs on both HPDv3 and PickScore, we run the iterative pipeline of Section~\ref{sec:procedure} for $R{=}10$ rounds on $|\mathcal{D}_{\text{train}}|{=}256$ preference pairs and retain the Top-$N$ rules with $N{=}20$. Each round mines $16$ hard pairs and discards any pair that has been selected more than $4$ times across rounds (stale-pair cap), preventing degenerate loops on label noise. Validation uses the same source corpus's held-out split.

\paragraph{Sparse rubric selection.}
The $\ell_1$-regularized logistic regression in Eq.~\eqref{eq:lr_bb} is fit with scikit-learn's \texttt{liblinear} solver: penalty \texttt{l1}, regularization strength $C{=}1.0$, no intercept, and a fixed random state of $42$. Coefficients with magnitude below $10^{-4}$ are treated as zero; positive coefficients are sorted in descending order and the top $N{=}20$ define $\mathcal{R}^{(t)}_{\text{retained}}$.

\paragraph{Curriculum-bucketed hard-pair mining.}
Misranked pairs under the current retained rubric set are partitioned into the three buckets defined in Section~\ref{sec:procedure}---\emph{small-margin}, \emph{large-margin}, and \emph{high-reward wrong}---using a margin percentile of $0.3$ and a reward quantile of $0.7$. The $16$ hard pairs sampled in round $r$ are drawn from these buckets with phase-dependent weights $(w_{\text{small}}, w_{\text{large}}, w_{\text{high}})$:
\begin{equation*}
(w_{\text{small}}, w_{\text{large}}, w_{\text{high}}) =
\begin{cases}
(0.6,\, 0.4,\, 0.0) & r < 3 \quad \text{(early)} \\
(0.5,\, 0.3,\, 0.2) & 3 \le r < R{-}1 \quad \text{(mid)} \\
(0.3,\, 0.3,\, 0.4) & r \ge R{-}1 \quad \text{(late)}.
\end{cases}
\end{equation*}
Early rounds emphasize large-margin errors that expose missing preference dimensions; later rounds shift mass to high-reward wrong cases that drive finer-grained rubrics for distinguishing strong generations.

\paragraph{Summary table.}
Table~\ref{tab:hparams_pipeline} consolidates the values above for reproducibility.

\begin{table*}[htbp]
\centering
\small
\begin{tabular}{llc}
\toprule
\textbf{Component} & \textbf{Hyperparameter} & \textbf{Value} \\
\midrule
\multirow{2}{*}{Models}
 & VLM judge backbone               & Qwen3-VL-8B-Instruct \\
 & Rule generator                   & Gemini-3-Flash \\
\midrule
\multirow{4}{*}{Sampling}
 & VLM judge temperature            & $0.0$ \\
 & VLM judge max output tokens      & $16$ \\
 & Rule-generator temperature       & $0.1$ \\
 & Rule-generator thinking budget   & $1024$ \\
\midrule
\multirow{4}{*}{Refinement loop}
 & \# rounds $R$                    & $10$ \\
 & \# seed pairs $|\mathcal{D}_{\text{train}}|$ & $256$ \\
 & Retained rules per round (Top-$N$) & $20$ \\
 & Hard pairs per round             & $16$ \\
\midrule
\multirow{5}{*}{$\ell_1$ logistic regression}
 & Solver                            & liblinear \\
 & Regularization strength $C$       & $1.0$ \\
 & Intercept                         & disabled \\
 & Non-zero coefficient threshold    & $10^{-4}$ \\
\midrule
\multirow{3}{*}{Hard-pair mining}
 & Margin percentile (small/large split) & $0.3$ \\
 & Reward quantile (high-reward bucket)  & $0.7$ \\
 & Stale-pair cap (max repeats)          & $4$ \\
\bottomrule
\end{tabular}
\caption{\small \textbf{Hyperparameters of the AutoRubric-T2I pipeline} used to produce the main results on HPDv3 and PickScore. Values are shared across both source corpora.}
\label{tab:hparams_pipeline}
\end{table*}

\vspace{-4mm}
\section{Qualitative Examples}
\label{app:qualitative_rl}
\begin{figure*}[htbp]
    \centering
    \includegraphics[width=0.8\textwidth]{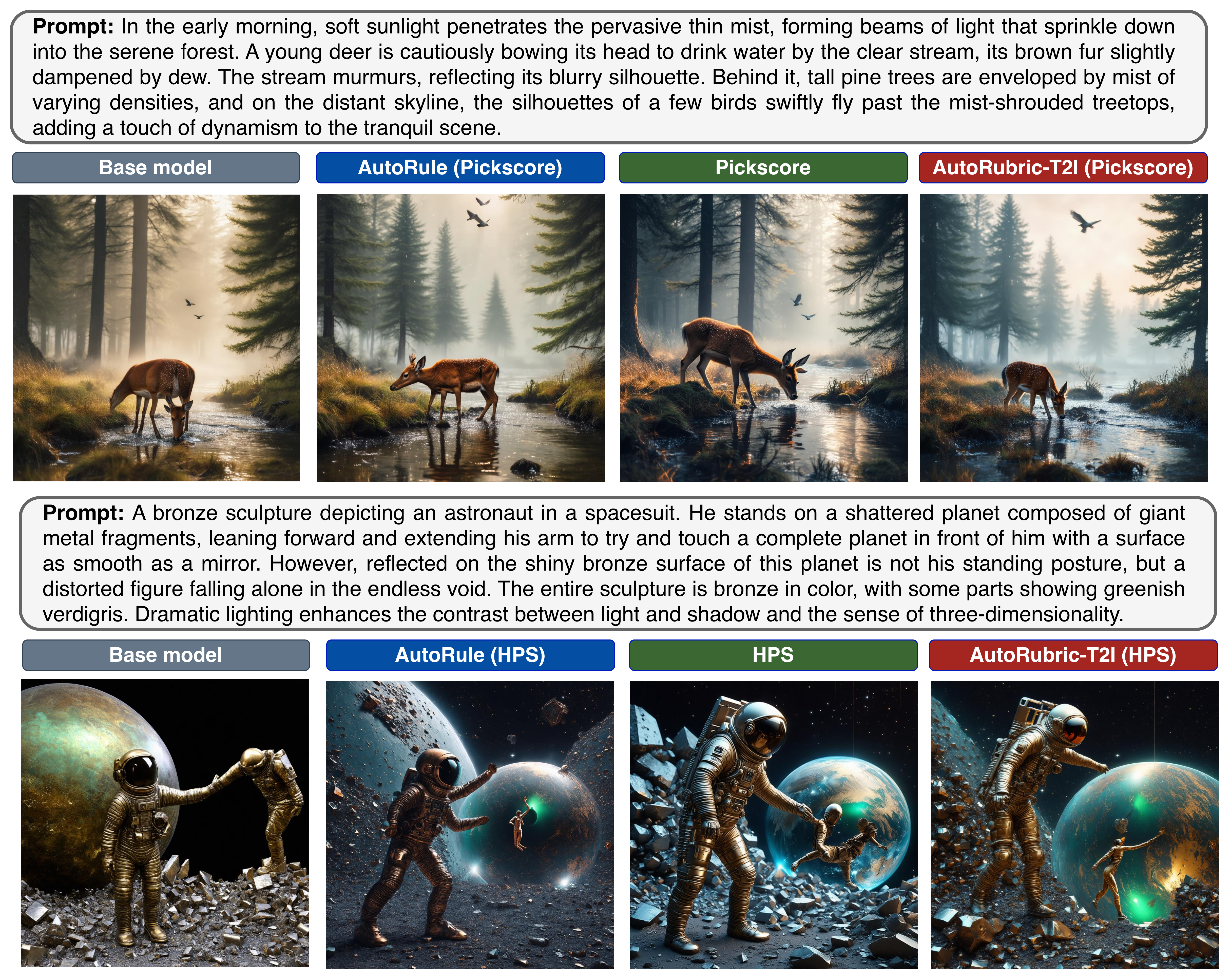}
    \caption{\small Qualitative comparison of downstream T2I RL policies. AutoRubric-T2I better preserves prompt-specific objects, relations, and fine-grained details compared with the base model, scalar reward optimization, and AutoRule-based rubric rewards.
}
\label{fig:qualitative_rl}
\end{figure*}

\begin{figure*}[htbp]
\centering
\includegraphics[width=0.9\textwidth,height=0.9\textheight,keepaspectratio]{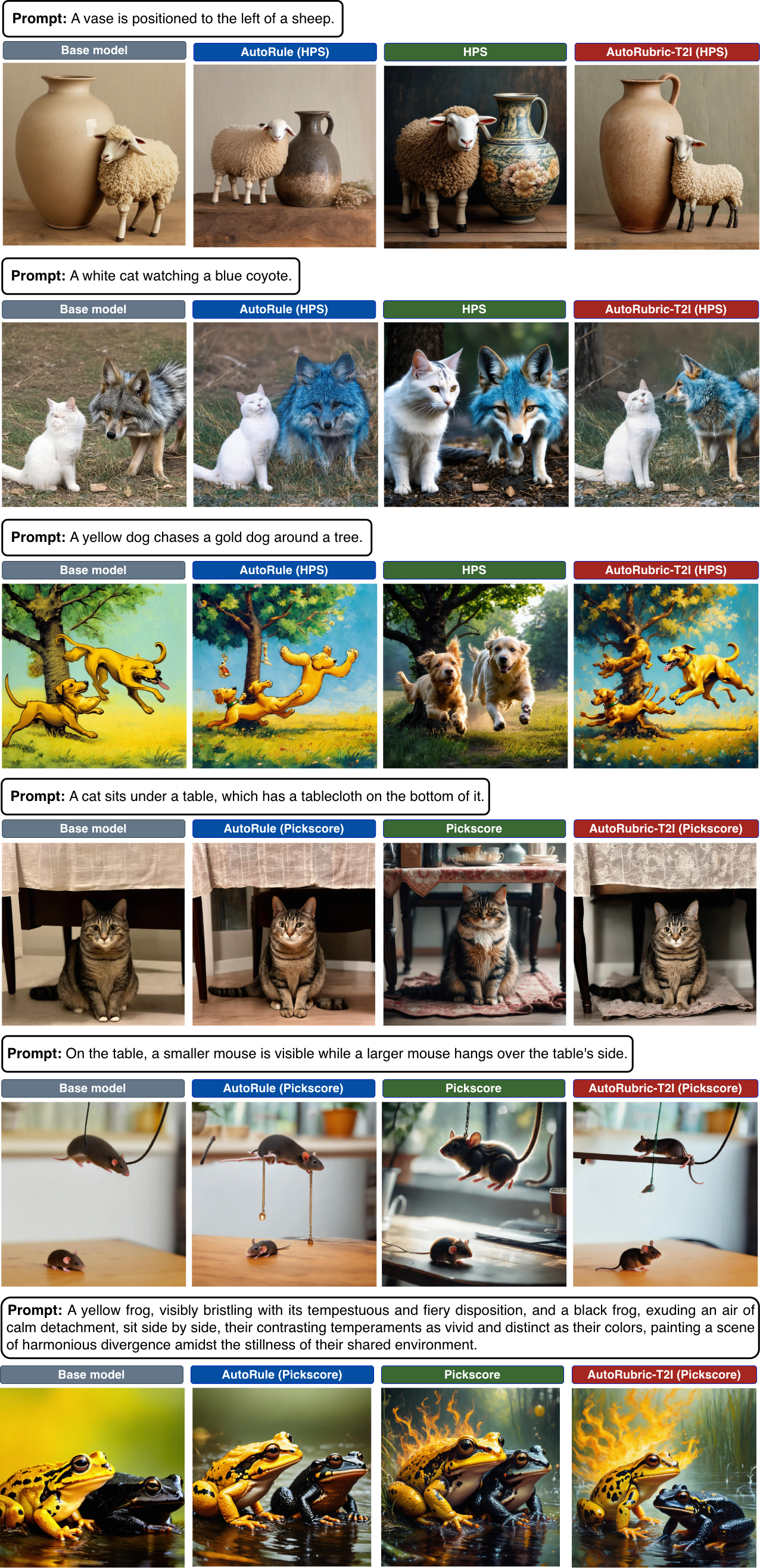}
\caption{\small Qualitative examples from downstream RL fine-tuning with AutoRubric-T2I rewards. Each row shows a text prompt and the corresponding generated image, demonstrating improved prompt alignment, object placement, attribute accuracy, and overall visual quality after RL training.}
\label{fig:rl_examples}
\end{figure*}

\vspace{-2mm}
We present qualitative examples of images generated by models fine-tuned with AutoRubric-T2I rewards via the Flow-GRPO pipeline. Figure~\ref{fig:qualitative_rl} and ~\ref{fig:rl_examples} show representative prompt--image pairs from UniGenBench++ and TIIF, illustrating improvements in prompt faithfulness, compositional accuracy, and visual quality compared to the base model.

\vspace{-4mm}
\section{Runtime and Data-needed Analysis}
\label{app:cost_comparison}
\vspace{-2mm}
A key motivation of \textbf{AutoRubric-T2I} is to reduce the data and training cost of conventional reward modeling. CLIP-based reward models such as HPSv2~\cite{wu2023hpsv2} and PickScore~\cite{kirstain2023pick} require 137K--798K human preference pairs and multi-GPU training. Recent VLM-based reward models~\cite{wang2025unified} reduce the annotation requirement but still require gradient-based training on 10K--100K+ preference pairs. In contrast, AutoRubric-T2I uses only \textbf{256 preference pairs} and requires \textbf{no neural reward model training}. Its main cost is a one-time rubric refinement stage.

\paragraph{One-time rubric refinement cost.}
Our final pipeline runs for 10 refinement rounds. In each round, we mine 16 hard preference pairs and use them to generate new candidate rubrics. The VLM scoring is performed with Qwen3-8B~\cite{yang2025qwen3} served by vLLM on 4$\times$ NVIDIA A6000 GPUs. Rule generation uses the Gemini API~\cite{gemini3_system_card} with \texttt{temperature=0.1} and \texttt{thinking\_budget=1024}. The $\ell_1$-regularized logistic regression for rubric selection runs on CPU and takes less than one second per round.

Table~\ref{tab:wallclock_setup} summarizes the wall-clock cost. Since the final setting keeps Top-20 rubrics and uses 10 rounds, the total number of VLM scoring calls is substantially smaller than earlier settings. The entire refinement stage finishes in approximately \textbf{2--4 hours on 4$\times$ A6000 GPUs}. This cost is incurred only once and the resulting weighted rubric set can be reused for downstream evaluation and RL.

\begin{table}[h]
\centering
\small
\begin{tabular}{lcr}
\toprule
\textbf{Component} & \textbf{Hardware} & \textbf{Est. Time} \\
\midrule
VLM scoring (${\sim}$180K calls) 
& Qwen3-8B / vLLM on 4$\times$ A6000 
& 1.5--3 hours \\
Validation scoring (${\sim}$20K--40K calls) 
& Qwen3-8B / vLLM on 4$\times$ A6000 
& 15--30 min \\
Rule generation (320 calls) 
& Gemini API 
& ${\sim}$15--30 min \\
Logistic regression (10 fits) 
& CPU 
& $<$1 min \\
\midrule
\textbf{Total} & & \textbf{2--4 hours} \\
\bottomrule
\end{tabular}
\vspace{2mm}
\caption{\small \textbf{Wall-clock time breakdown for AutoRubric-T2I rubric refinement} under the final setting: 10 rounds, 256 preference pairs, Top-20 rubrics, and 16 hard pairs per round.}
\label{tab:wallclock_setup}
\end{table}

\vspace{-8mm}
\paragraph{Cost during RL.}
At deployment time, each generated image is independently scored by the 20 selected rubrics, which requires 20 VLM forward passes. For a representative RL setting with 10K prompts and 4 rollouts per prompt, this corresponds to 40K generated images and 800K rubric-scoring calls. These calls are parallel across images and rubrics, so throughput scales directly with additional vLLM replicas.

\begin{table}[htbp]
\centering
\resizebox{\textwidth}{!}{%
\begin{tabular}{lcccccc}
\toprule
\textbf{Method} 
& \textbf{\#Pref. Pairs} 
& \textbf{Training Cost} 
& \textbf{Calls / Image} 
& \textbf{FLOPs / Image} 
& \textbf{Open-Source?} 
& \textbf{Interpretable?} \\
\midrule
CLIP-based RM 
& 137K--798K 
& 8--32 GPUs $\times$ days 
& 1 
& ${\sim}1.5 \times 10^{11}$ 
& Yes 
& No \\
VLM-based RM 
& 10K--100K+ 
& 8--64 GPUs $\times$ hrs--days 
& 1 
& ${\sim}2.4 \times 10^{13}$ 
& Varies 
& No \\
Zero-shot VLM Judge 
& 0 
& None 
& 1 
& ${\sim}2.6 \times 10^{13}$ 
& Yes 
& Partial \\
RubricRL 
& 0 
& None 
& 10 
& API 
& No 
& Yes, uniform \\
\textbf{AutoRubric-T2I} 
& \textbf{256} 
& \textbf{2--4 hrs on 4$\times$ A6000} 
& \textbf{20} 
& $\boldsymbol{{\sim}5.1 \times 10^{14}}$ 
& \textbf{Yes} 
& \textbf{Yes, learned weights} \\
\bottomrule
\end{tabular}%
}
\vspace{2mm}
\caption{\small \textbf{End-to-end cost comparison.} AutoRubric-T2I trades higher inference cost for substantially lower annotation cost, no gradient-based reward-model training, and interpretable per-rubric scores with learned weights. FLOPs are estimated by $2P \times T$ for open-weight models.}
\label{tab:cost_comparison}
\end{table}

\vspace{-6mm}
\paragraph{Summary.}
AutoRubric-T2I shifts the cost of reward modeling from large-scale annotation and reward-model training to a lightweight, one-time rubric refinement procedure. It requires only 256 preference pairs, runs in a few hours on 4$\times$ A6000 GPUs, and produces an interpretable weighted rubric set. Although per-image inference is more expensive than scalar reward models, the cost enables fine-grained diagnostics, rubric-level reward shaping, and improved robustness against scalar reward hacking.

\begin{figure*}[htbp]
    \centering
    \includegraphics[width=0.95\textwidth]{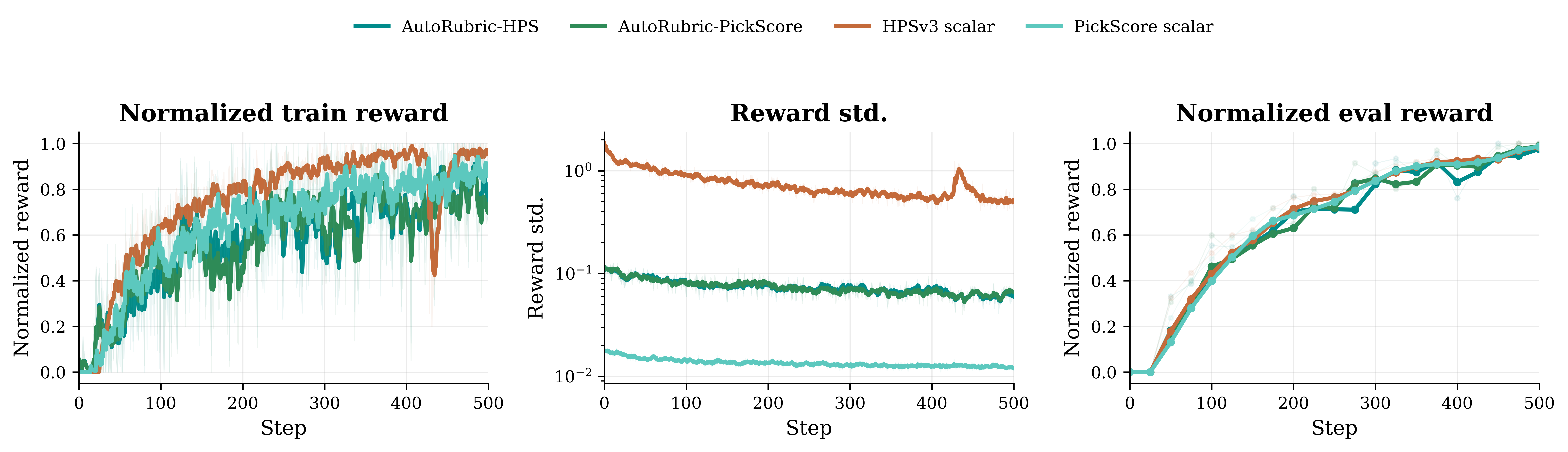}
    \caption{\small Training dynamics of scalar and rubric-based T2I rewards.}
    \label{fig:training_curves}
\end{figure*}
\vspace{-4mm}

\begin{figure*}[htbp]
    \centering
    \includegraphics[width=0.85\textwidth]{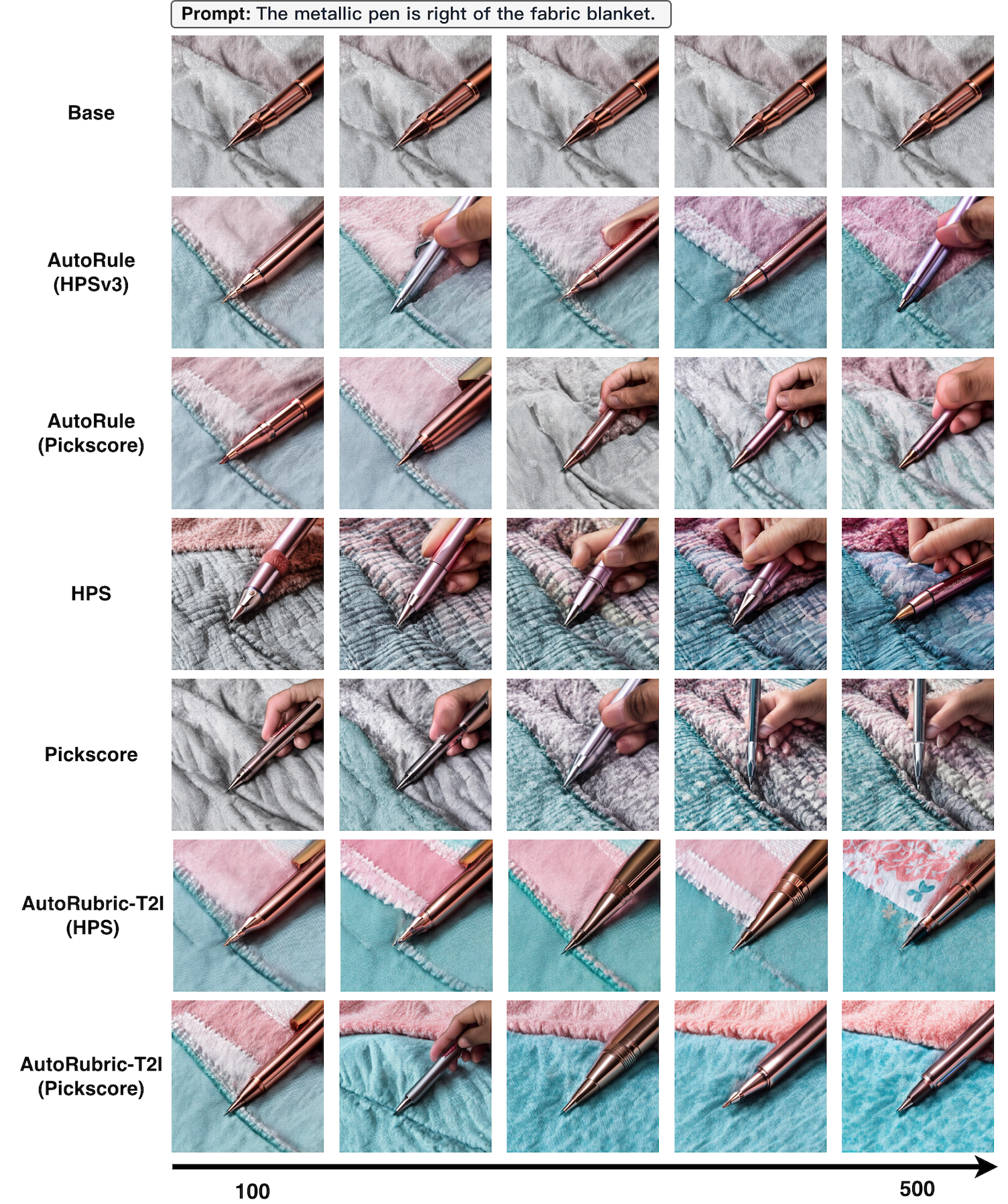}
    \caption{\small The evolution of generation quality of RL using AutoRubrics and other scalar reward models. The visual quality of scalar reward models degrades notably while the reward increases.
}
\label{fig:reward_hacking_rl}
\end{figure*}

\section{Training Dynamics}
\label{app:training_dynamics}
\vspace{-2mm}
We provide training curves of T2I-RL referenced in Section~\ref{sec:reward_hacking}. In Figure~\ref{fig:training_curves}, we visualize normalized training and evaluation rewards together with reward standard deviation. AutoRubric-T2I achieves steadily improving training and evaluation rewards while maintaining substantially lower reward variance. In contrast, HPSv3 exhibits much larger reward dispersion throughout training, suggesting a noisier optimization signal, whereas PickScore has extremely low reward variance, indicating limited sensitivity for distinguishing high-capability generations. Qualitative results throughout RL training are visualized in Figure~\ref{fig:reward_hacking_rl}, where scalar reward shows clear visual evidence of reward hacking.

\vspace{-2mm}
\section{Limitations and Broader Impact}
\label{app:limitations}
 
\paragraph{Domain specificity of learned weights.}
The $\ell_1$-regularized weights are fit to the preference distribution of the training corpus (e.g., HPSv3 or PickScore), so the rubric \emph{texts} remain broadly applicable, but their relative importance may shift on out-of-domain prompts such as typography-heavy or highly stylized images. Because rubric refinement requires only 256 pairs, the natural remedy is to re-fit the logistic regression on a small in-domain sample with the rubric set fixed, or to learn prompt-conditional weights.

\paragraph{RL training integration.}
AutoRubric-T2I can be directly integrated into Flow-GRPO as a drop-in reward by using the learned weighted sum of per-rubric scores. Beyond this simple scalarization, the explicit per-rubric decomposition provides a promising opportunity for rubric-aware advantage estimation, enabling finer-grained credit assignment during T2I reinforcement learning. However, how to fully exploit both rubric-level rewards and conventional scalar rewards for more effective advantage estimation remains an important direction for future work.

\paragraph{Broader impact.}
By replacing opaque scalar reward models with human-readable rubrics and learned weights, AutoRubric-T2I makes the criteria that drive T2I optimization inspectable and auditable, which helps diagnose and correct biases (e.g., systematic under-weighting of culturally diverse aesthetics) directly from the weights. The low data requirement and open-source infrastructure further lower the barrier for groups without large annotation pipelines or proprietary APIs, and we do not foresee negative societal impacts beyond those inherent to T2I generation broadly.

\section{Prompt}
\label{app:prompts}

\begin{figure*}[htbp]
\centering
\includegraphics[width=0.85\textwidth]{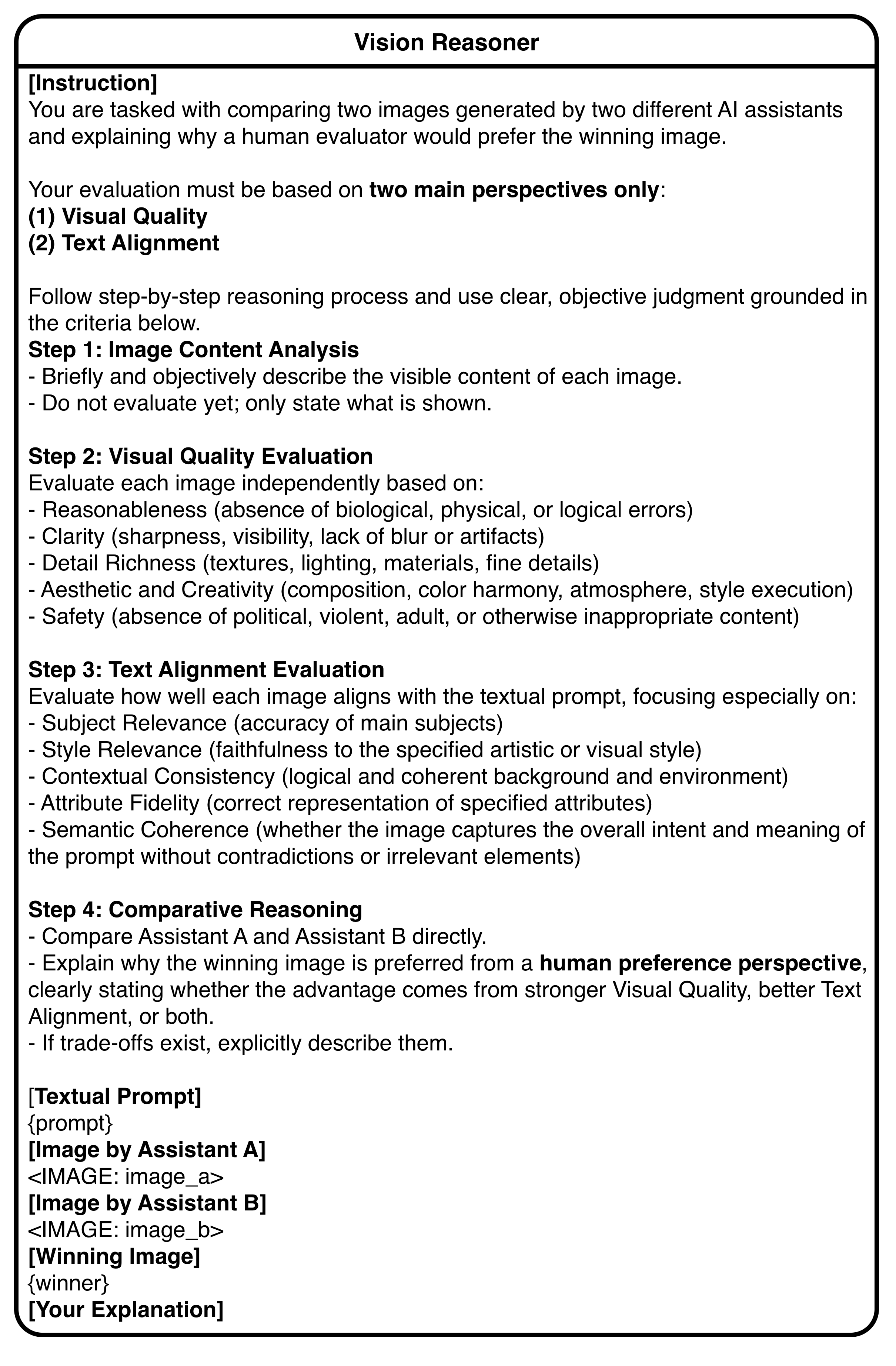}
\caption{\small Seed rubric generation, stage 1: vision reasoner that produces a step-by-step preference rationale for each image pair.}
\label{fig:prompt_seed}
\end{figure*}

\begin{figure*}[htbp]
\centering
\includegraphics[width=0.8\textwidth]{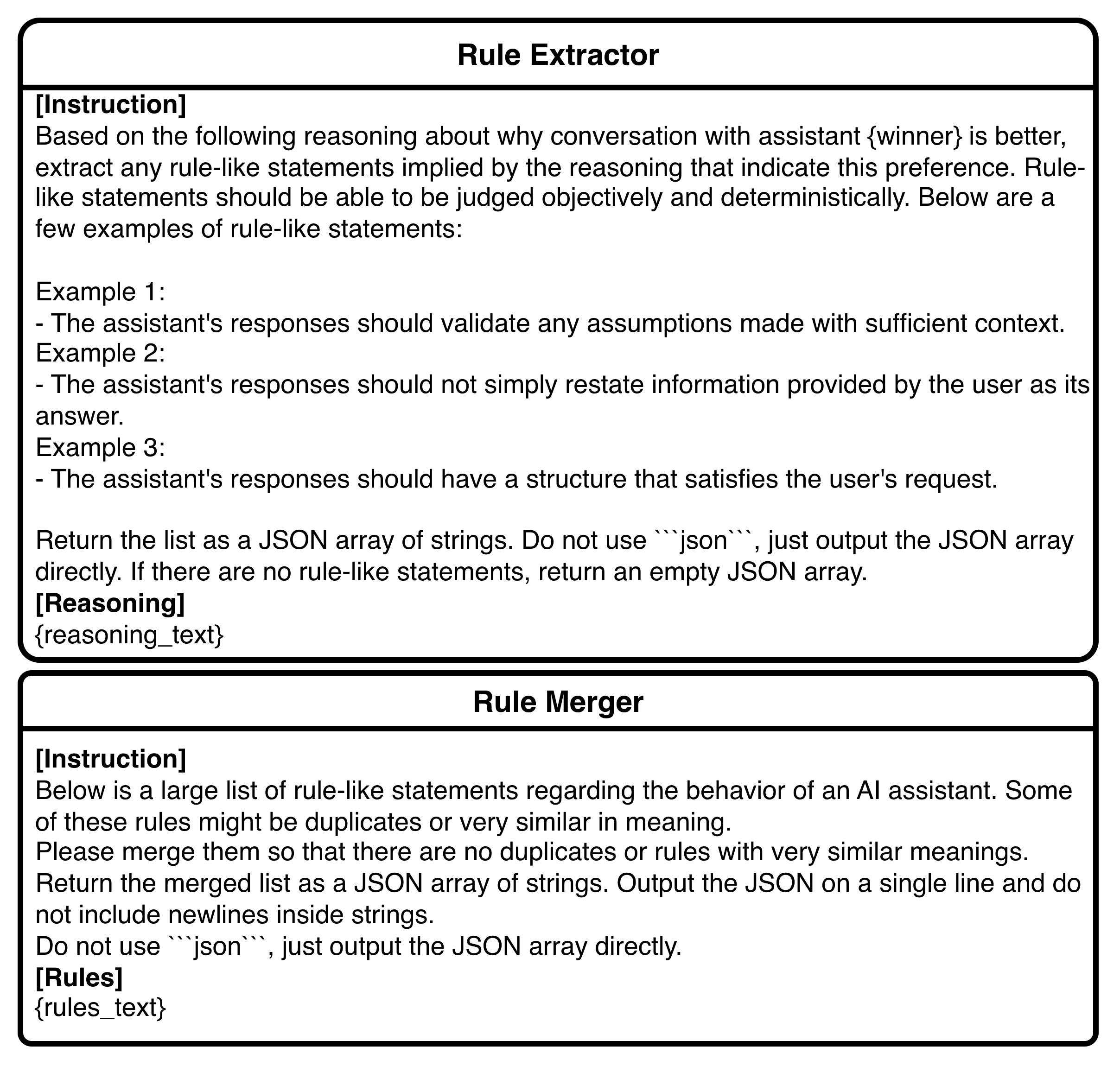}
\caption{\small Seed rubric generation, stages 2-3: rule extractor rule merger.}
\label{fig:prompt_extract_merge}
\end{figure*}

\begin{figure*}[htbp]
\centering
\includegraphics[width=0.8\textwidth]{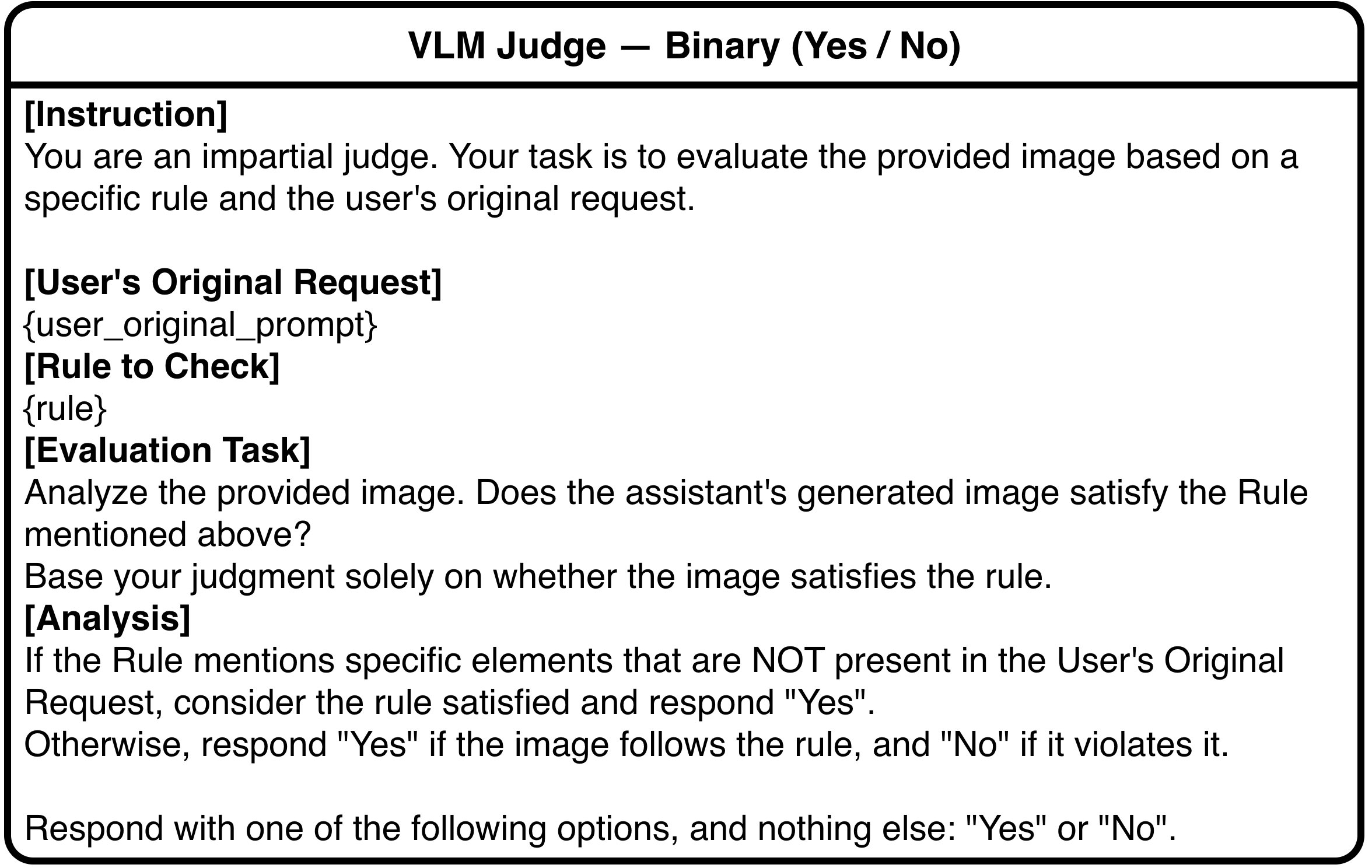}
\caption{\small VLM judge templates: Yes/No binary scoring.}
\label{fig:prompt_judges}
\end{figure*}

\begin{figure*}[htbp]
\centering
\includegraphics[width=0.85\textwidth]{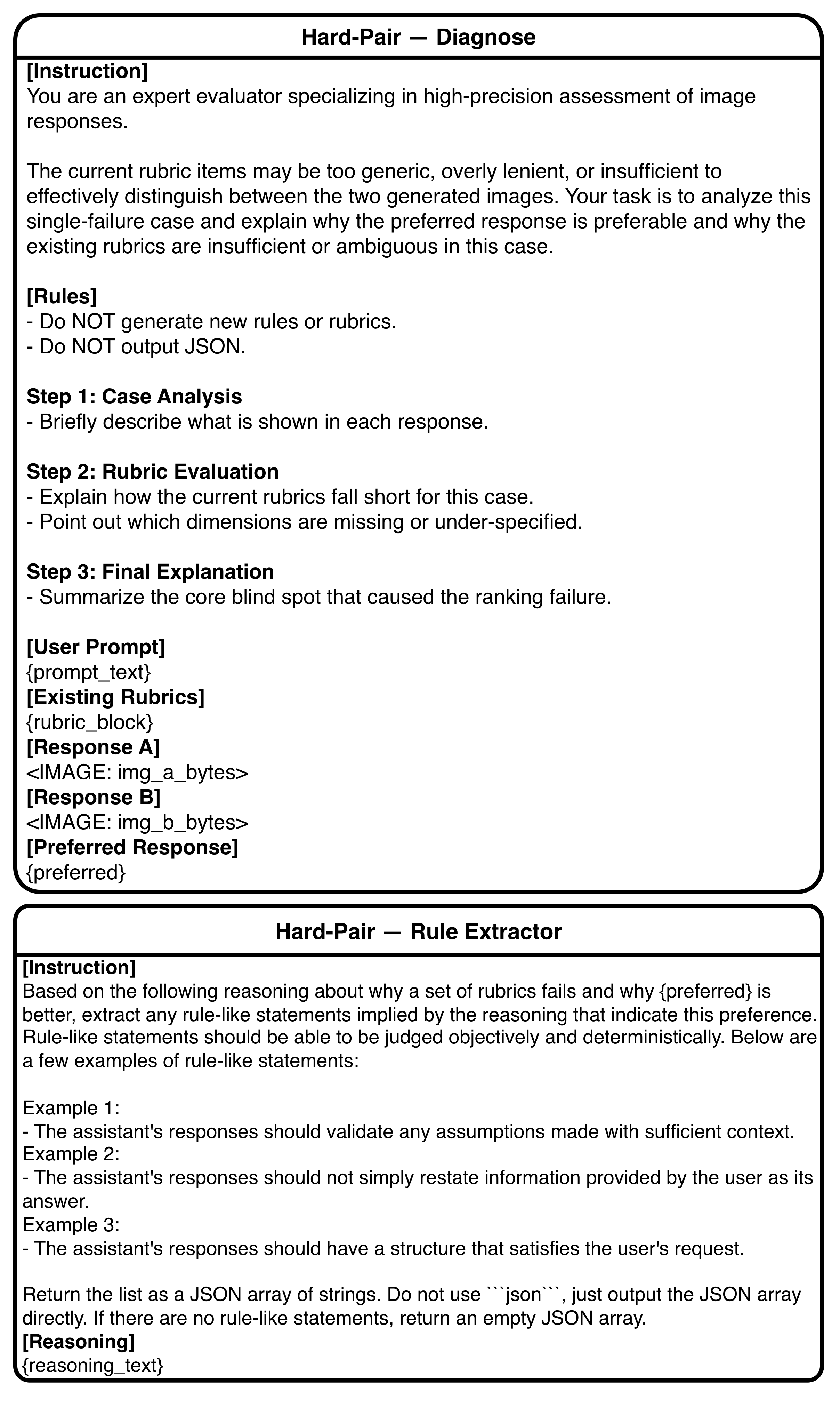}
\caption{\small Hard-pair refinement prompt.}
\label{fig:prompt_hardpair}
\end{figure*}

Figures~\ref{fig:prompt_seed}--\ref{fig:prompt_hardpair} present the complete set of prompt templates used at all stages of the AutoRubric-T2I pipeline: vision reasoner and rule extractor/merger for seed rubric generation, VLM judge templates, and hard-pair diagnosis and rule extraction prompts used during iterative refinement.

\FloatBarrier

\FloatBarrier

\section{Quantitative Human Evaluation}
\label{app:user_survey}

To supplement the quantitative human evaluation in Section~\ref{sec:human_eval}, we provide a screenshot of the user survey interface used in our study. The survey was conducted with 30 graduate-student annotators. Each annotator was asked to evaluate 20 T2I prompts, resulting in 600 total human judgments.

For each question, annotators were shown a text prompt at the top, and four anonymized image candidates labeled A--D. These candidates correspond to outputs from the base model, scalar-reward optimization, AutoRule-based optimization, and AutoRubric-T2I, with the display order randomized to reduce positional bias. Annotators were instructed to select the single image that best satisfied the text prompt, considering both visual quality and prompt alignment. In particular, they were encouraged to focus on whether the generated image correctly captured the requested objects, spatial relations, attributes, scene layout, and other fine-grained constraints in the prompt.

\begin{figure}[h]
    \centering
    \includegraphics[width=0.6\linewidth]{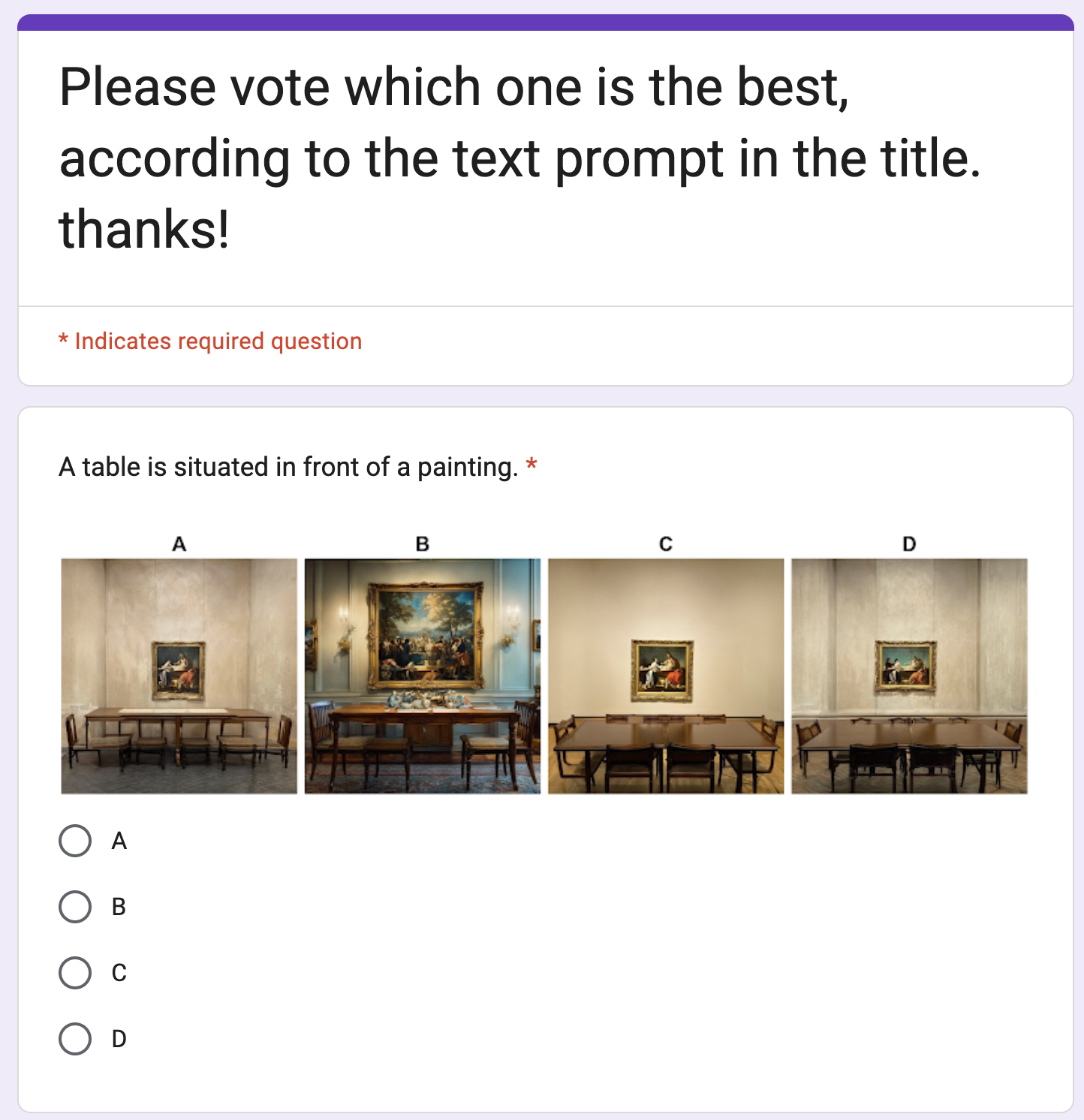}
    \caption{Screenshot of the human evaluation survey interface. Annotators were asked to choose the best image according to the text prompt shown in the question title.}
    \label{fig:user_survey_screenshot}
\end{figure}

Figure~\ref{fig:user_survey_screenshot} shows an example survey question. The instruction explicitly asks users to vote for the best image according to the text prompt, and each question requires exactly one choice among the four candidates. This simple forced-choice design avoids requiring annotators to assign calibrated numerical scores, while still providing a direct measure of human preference among RL policies.

\vspace{-4mm}
\section{Optimized Rubric Set}
\label{app:optimized_rubrics}
\vspace{-2mm}
This section reports the final weighted rubric sets produced by the AutoRubric-T2I pipeline for each (VLM judge, source corpus) configuration. Rules are ranked in decreasing order of their fitted $\ell_1$-regularized logistic-regression weights ($\bar{w}$); rules with effectively zero weight are pruned from the displayed Top-$N$ set.

\FloatBarrier
\vspace{-4mm}


\begin{figure}[hbp]
\centering
\vspace{-2mm}
\includegraphics[width=0.88\linewidth]{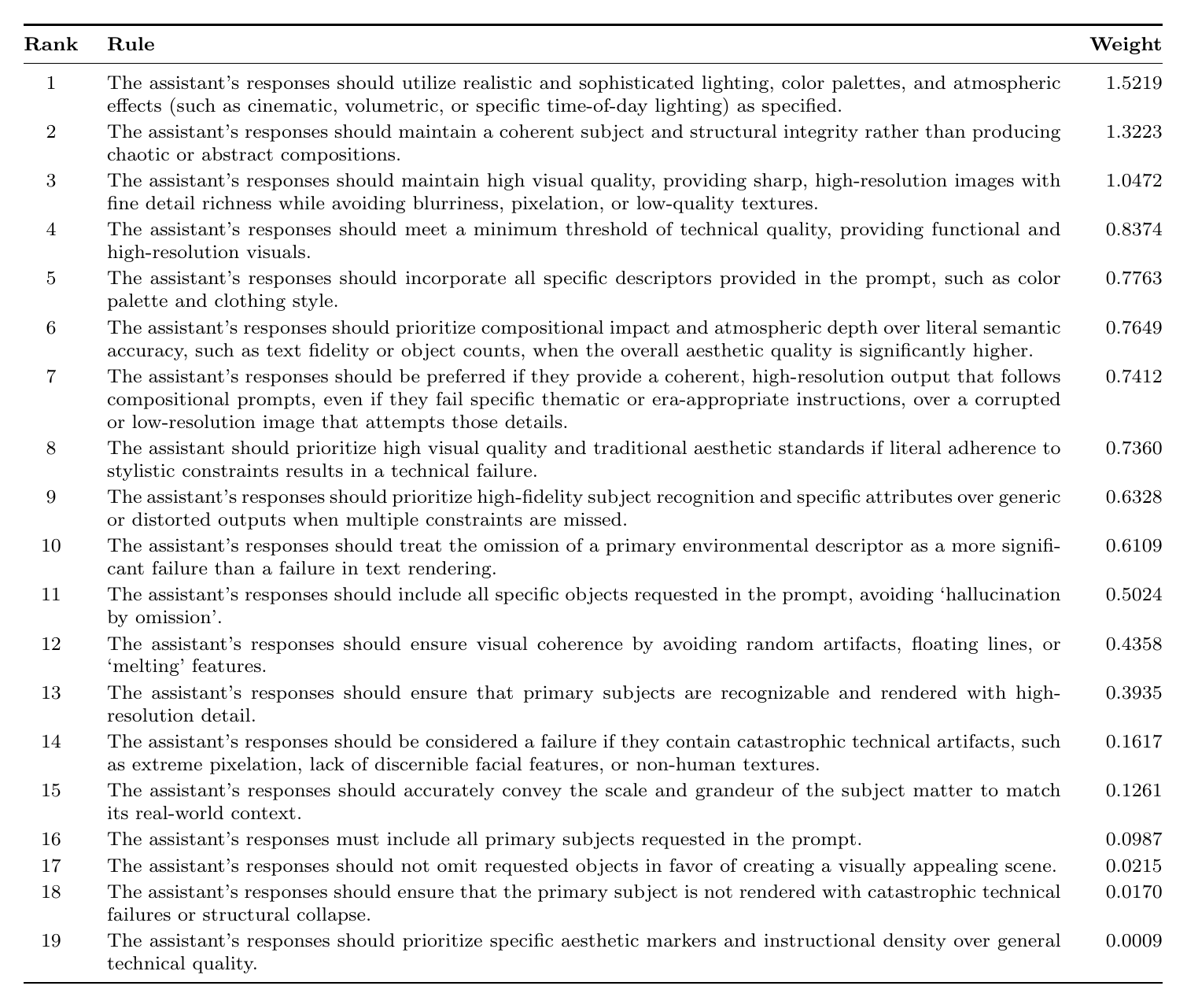}
\vspace{-2mm}
\caption{\small Optimized rubric set for Qwen-3-VL-8B trained on HPSv3 preference pairs (round 3).}
\label{fig:rubric_qwen8b_hpsv3}
\end{figure}

\begin{figure}[hbp]
\centering
\vspace{-4mm}
\includegraphics[width=0.88\linewidth]{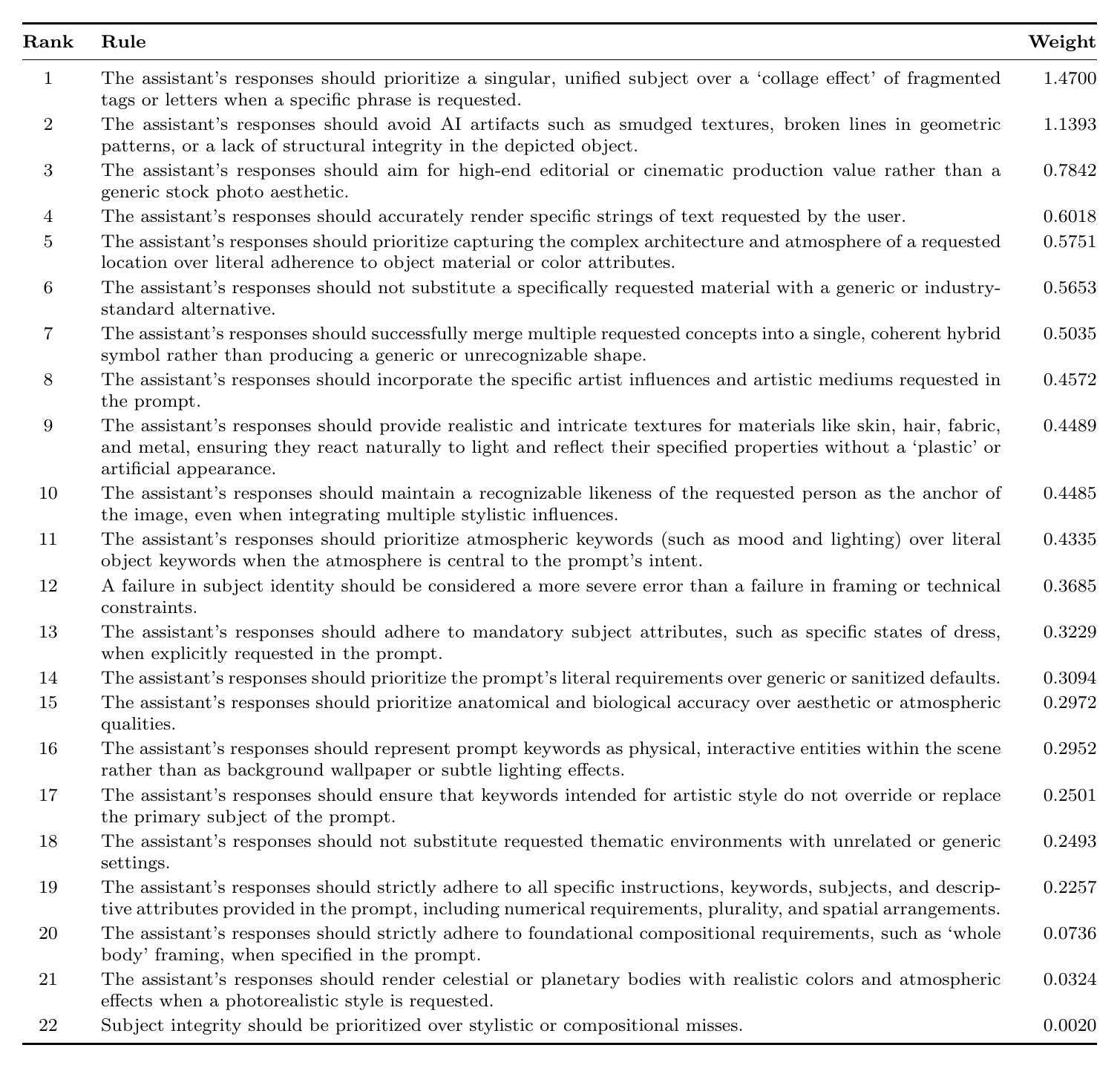}
\vspace{-2mm}
\caption{\small Optimized rubric set for Qwen-3-VL-8B trained on PickScore preference pairs (round 6).}
\label{fig:rubric_qwen8b_pickscore}
\end{figure}



\begin{figure}[hbp]
\centering
\includegraphics[width=0.98\linewidth]{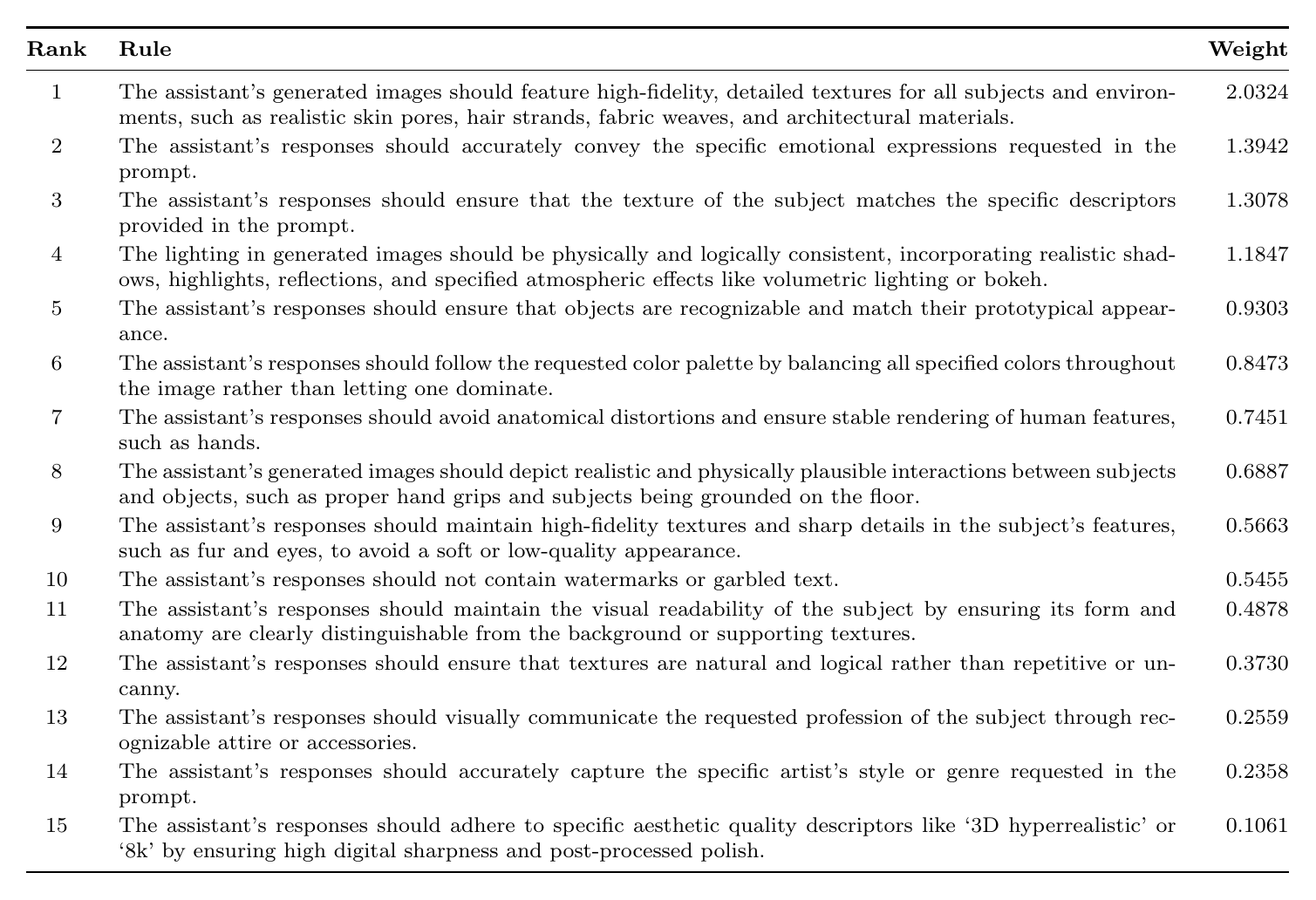}
\caption{\small Optimized rubric set for Qwen-3-VL-32B trained on HPSv3 preference pairs (round 3).}
\label{fig:rubric_qwen32b_hpsv3}
\end{figure}


\begin{figure}[hbp]
\centering
\includegraphics[width=0.98\linewidth]{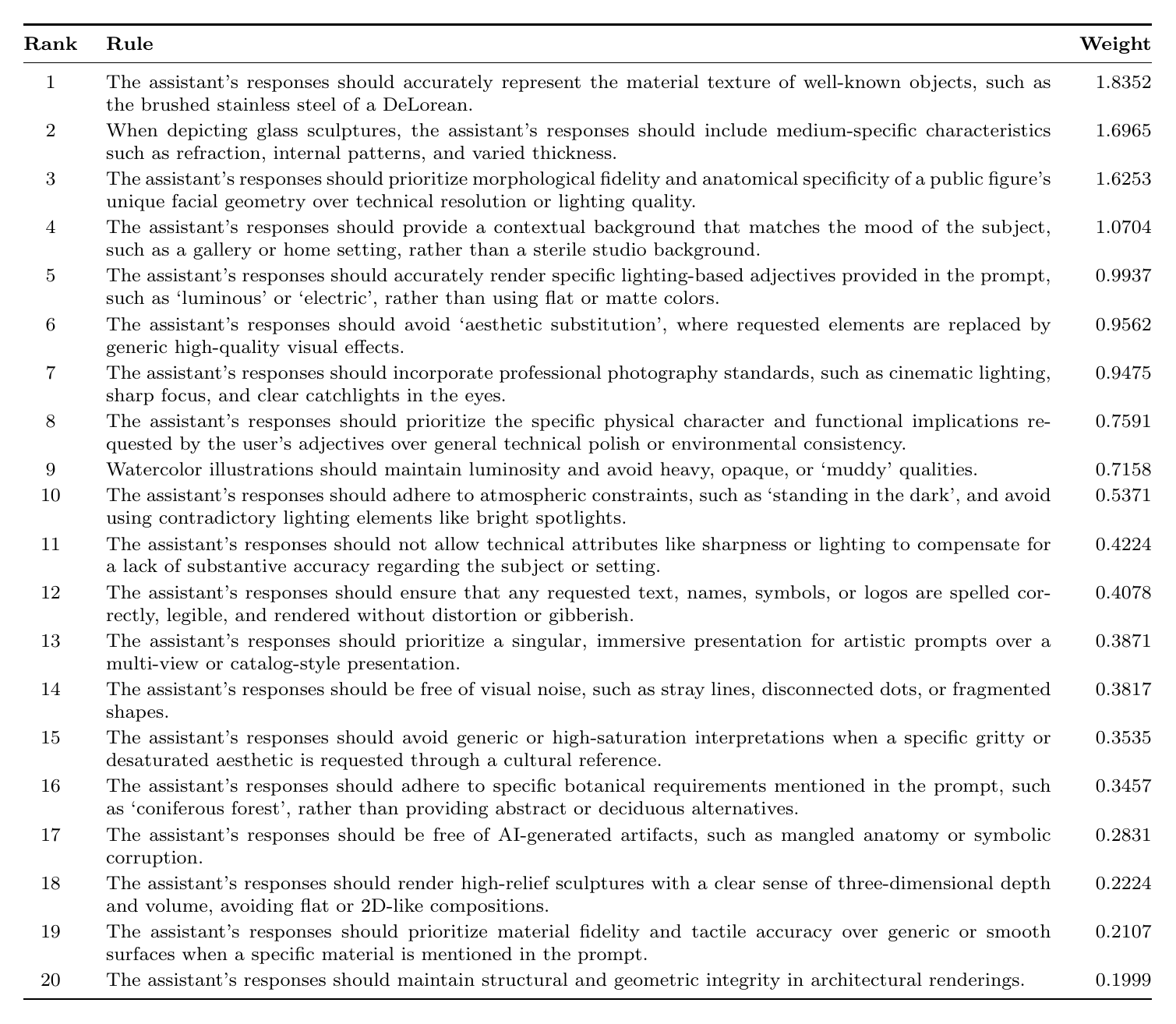}
\caption{\small Optimized rubric set for Qwen-3-VL-32B trained on PickScore preference pairs (round 6).}
\label{fig:rubric_qwen32b_pickscore}
\end{figure}

\FloatBarrier


\end{document}